\documentclass[final]{cvpr}

\usepackage{times}
\usepackage{epsfig}
\usepackage{graphicx}
\usepackage{amsmath}
\usepackage{amssymb}
\usepackage{float}
\usepackage{subfigure}
\usepackage{booktabs}
\usepackage[dvipsnames]{xcolor}

\usepackage[pagebackref=true,breaklinks=true,colorlinks,bookmarks=false]{hyperref}

\def\cvprPaperID{11046} % *** Enter the CVPR Paper ID here
\def\confYear{CVPR 2021}

\begin{document}

%%%%%%%%% TITLE
\title{CubifAE-3D: Monocular Camera Space Cubification for Auto-Encoder based 3D Object Detection}

\author{Shubham Shrivastava and Punarjay Chakravarty \\
Ford Greenfield Labs, Palo Alto\\
{\tt\small \{sshriva5,pchakra5\}@ford.com}}
% \and

\maketitle

% % **** Enter the paper title here

%\thispagestyle{empty}
%%%%%%%%% BODY TEXT - ENTER YOUR RESPONSE BELOW

\begin{abstract}
We introduce a method for 3D object detection using a single monocular image. Starting from a synthetic dataset, we pre-train an RGB-to-Depth Auto-Encoder (AE). The embedding learnt from this AE is then used to train a 3D Object Detector (3DOD) CNN which is used to regress the parameters of 3D object poses after the encoder from the AE generates a latent embedding from the RGB image. We show that we can pre-train the AE using paired RGB and depth images from simulation data once and subsequently only train the 3DOD network using real data, comprising of RGB images and 3D object pose labels (without the requirement of dense depth). Our 3DOD network utilizes a particular `cubification' of 3D space around the camera, where each cuboid is tasked with predicting N object poses, along with their class and confidence values. The AE pre-training and this method of dividing the 3D space around the camera into cuboids give our method its name - CubifAE-3D. We demonstrate results for monocular 3D object detection in the Autonomous 
Vehicle (AV) use-case with the Virtual KITTI 2 and the KITTI datasets.
\end{abstract}

\section{Introduction}

Detecting objects around the ego-vehicle is one of the fundamental tasks for the perception system on an AV. The perception of objects such as other vehicles, pedestrians, bicycles, and their relative positions in 3D space around the ego-vehicle are then used by the vehicle’s path planning system to chart a collision free route through it.

Detecting objects as 2D bounding boxes in monocular camera images is a relatively mature technology, and techniques like YOLO \cite{YOLO} work reasonably well in the real world. However, the same cannot be said for detecting these objects in 3D space using monocular cameras. Most methods for 3D object detection available today use high-cost sensors such as LIDAR \cite{zhou2019endtoend, wang2019frustum, shi2018pointrcnn, zhou2018voxelnet}. 

\begin{figure}[H]%[hpt!]
\centering
    % \begin{subfigure}
    % \centering
    % \includegraphics[width=0.95\linewidth]{fig/qual_res/intro_nuscenes.jpg}
    % \end{subfigure}
    \begin{subfigure}
    \centering
    \includegraphics[width=0.95\linewidth]{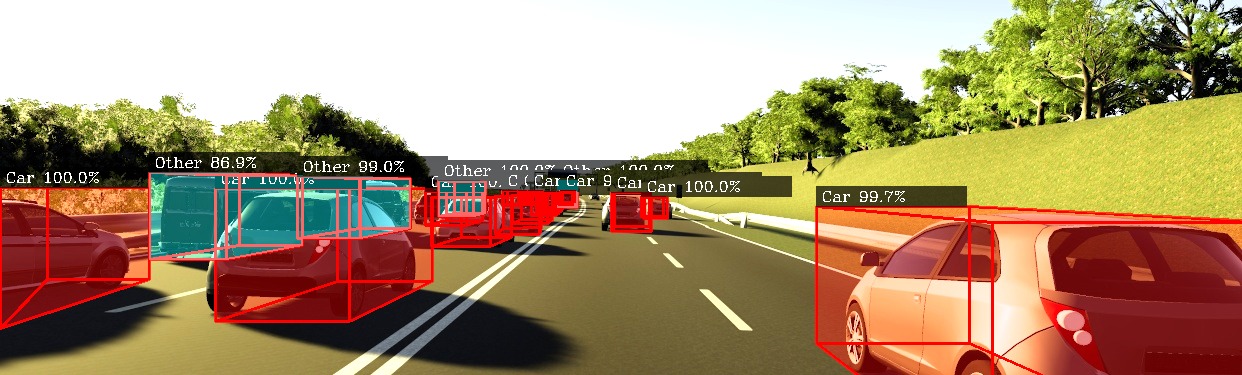}
    \end{subfigure} 
    \begin{subfigure}
    \centering
    \includegraphics[width=0.95\linewidth]{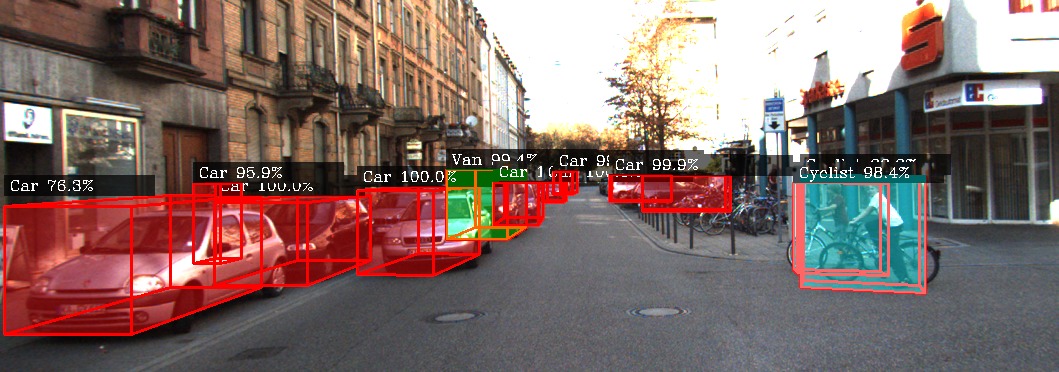}
    \end{subfigure}
\caption{Object pose predictions using \emph{CubifAE-3D} on monocular RGB camera images from the \emph{vKITTI2} (top) and \emph{KITTI} (bottom) datasets.
% , colour coded as:
% \emph{red: car, yellow: truck, green: van, blue: pedestrian, cyan: all other classes.}
\label{fig:intro_img}}
\vspace*{-0.1cm} 
\end{figure}

% Jay: Forcing figure to stay here with \usepackage{float} and \begin{figure}[H]
% See: https://tex.stackexchange.com/questions/241403/adding-figures-on-first-page-of-documentclass-article
% Otherwise this was being pushed to the 2nd page.

Some work has also suggested the use of depth-maps to generate pseudo-lidar representation, which can subsequently be used for 3D object detection using state-of-the-art (SoTA) LIDAR based object detection methods \cite{wang2019pseudo,you2019pseudo}. 
%Jay: changing LiDAR to LIDAR everywhere for consistency

Here, we present a method for performing 3D object detection using a single RGB camera during inference. We assume that a dense depth map (as obtained from stereo / RGB-D camera) is available during training. Sparse depth maps (as obtained from LIDAR) could also be used for training by first transforming it into a dense representation \cite{ma2017sparsetodense, xu2019depth}. We first pre-train an RGB-to-depth auto-encoder in a fully-supervised manner and let the latent space learn an RGB-to-depth embedding, after which, the Decoder is disconnected and the output of the latent space is fed to the 3D Object Detection (3DOD) network, which is trained to predict 3D object poses. We call our model CubifAE-3D, \emph{pronounced Cubify-3D}. The first part of the name refers to the \emph{cubification} or voxellization of monocular camera space as a pre-processing step, and the AE refers to the Auto-Encoding of RGB-to-depth space. 

Furthermore, we show that the RGB-to-depth auto-encoder can to be pre-trained using \emph{simulation} data. This pre-trained latent encoding of the RGB image can subsequently be used to train the 3DOD network on a separate real dataset. The only annotation needed for the real dataset therefore, is the object pose itself corresponding to each camera frame. We do not use LIDAR or stereo data for training the network on the real dataset (though these sensing sources might be required to annotate the 3D labels themselves).
% and no other source of data (LIDAR, stereo camera pair) is required. 
Figure \ref{fig:intro_img} illustrates the 3D object pose predictions from a single RGB image made by our model on the \emph{vKITTI2} \cite{cabon2020virtual} and \emph{KITTI} \cite{Geiger2012CVPR} datasets. 

%Jay: We must emphasize that our method works on vehicles (rigid bodies) as well as pedestrians. Most other methods only detect vehicles.

% Top part of the image shows bounding-boxes drawn based on 2D projections of the 3D pose predictions, and the bottom part shows a birds-eye view of the same scene with the ego vehicle positioned at the center of the blue circle shown on the left looking towards the right of the image.
%Jay removed from here: TO Add to figures later on in the paper.

\section{Related Work}

Due to the expense of LIDAR scanners and the simultaneous advances in Deep Learning, camera-only techniques for 3D object detection have gained currency over the past few years. Pseudo-LIDAR \cite{wang2019pseudo,you2019pseudo} demonstrated that they could retrieve LIDAR-like point clouds from cameras. Camera images (monocular and stereo) are converted into a depth map, which is unprojected into the 3D camera space to get a point-cloud. 3D object poses are subsequently detected directly in the point cloud using a Point-net-like architecture \cite{qi2017pointnet} or converted to a Bird's Eye View (BEV) image, before being fed to a CNN for detecting the 3D pose of objects.  The BEV projection has the advantage of preserving object size and scale throughout the representation and is faster to process, allowing real-time detections. The BEV can be treated as an image, that can be scanned using 2D convolutions that are faster to compute compared to the 3D convolutions required in the point cloud generated from stereo.

% using an AVOD-like architecture \cite{ku2018joint}. Pseudo-LIDAR++ \cite{you2019pseudo} improves on the 3D pose estimation accuracy by adding a very sparse (4-beam) LIDAR to constrain the errors in the point-cloud generated by stereo depth map unprojection.

% Geometric constraints
Determining the 6-Dof pose of an object from a single 2D image is essentially an ill-posed problem, because the camera projection equation squashes the depth dimension so that a 2D point in an image corresponds to a ray in 3D space, and the 3D world point corresponding to its projected 2D point could lie along any point along that ray. Approaches like Mousavian et. al. \cite{mousavian20173d} and MonoGRNet \cite{qin2019monogrnet} combine Deep Learning with the geometric properties for the rigid bodies of vehicles to constrain this problem. 
% Deep Learnt 2D detectors are combined with an optimization approach during inference for 3D bounding box detection from monocular images. 
Mousavian et. al. \cite{mousavian20173d} uses Deep Learning to predict the orientation angle and the dimensions of the 3D bounding box corresponding to a 2D detection. These are then used to instantiate an oriented and sized 3D box in 3D space, along the ray corresponding to the centre of the 2D detection, and the depth (distance along the ray) is optimized by minimizing the 2D projection error of the instantiated 3D box relative to the 2D detection.

% is then moved to the right depth along the detection ray (after rotating it with the appropriate orientation angle)

% oordinate frame by minimizing its projection error compared to the 2D detection.
% They use the projection of putative 3D bounding box detections for a vehicle into 2D and use the error between this and the 2D detection of the vehicle to optimize for the fine-grained pose of the vehicle.
% determine the 3D pose of the vehicle using a combination of Deep Learning and optimization techniques.

% Part-based detectors:
Deep MANTA \cite{chabot2017deep} and Mono3D++ \cite{he2019mono3d++} use 2D vehicle part detectors in monocular images to further constrain the determination of their 3D poses using optimization. Deep-learnt vehicle part detectors for the location of vehicle parts in the 2D image are combined with the actual relative 3D locations of these parts to determine the pose of the vehicle in the camera coordinate frame. Mono3D++ \cite{he2019mono3d++} adds additional priors like monocular depth estimation and ground plane detection to the 3D part locations to improve performance.

% in addition to 3D part locations, adds additional priors like depth (from monocular depth estimation) and ground plane detection. 

% and use the actual 3D locations of these parts relative to the vehicle coordinate frame (learnt offline) and a 3D-to-2D matching technique like the Perspective-n-Point Projection algorithm \cite{lepetit2009epnp} to determine the pose of the vehicle in the camera coordinate frame. Mono3D++ \cite{he2019mono3d++}, in addition to 3D part locations, adds additional priors like depth (from monocular depth estimation) and ground plane detection. Conv-nets are used to predict a coarse orientation and scale of the 3D bounding box (corresponding to its 2D detection), in addition to detecting 2D part-landmarks. These are then combined with the 3D wireframe model of the car along with the depth and ground-plane priors to get the final 3D pose of the vehicle in an energy minimization framework.
% that is solved using the Ceres solver \cite{ceres-solver}.

An efficient way of training an object detector is to divide the image into a grid and learning anchors that are typical of the sizes and aspect ratios of the bounding boxes normally detected in that part of the image. Each grid cell is associated with a fixed number of anchors, each with a prototypical size and aspect ratio, where a training sample is associated with an anchor based on an intersection-over-union (IoU) score. The network is trained to output an offset from the anchor in terms of centre and dimensions of the ground truth bounding box relative to the anchor. This is the approach adopted by YOLO \cite{YOLO} for 2D object detection and more recently, by M3D-RPN \cite{brazil2019m3d}, a method that extends the anchor box idea to 3D space for monocular 3D object detection. M3D-RPN learns both 2D and 3D bounding box priors for anchors from the data and the network is trained to output 3D bounding box parameters that are offsets from these anchors.

Our \emph{CubifAE-3D} network takes inspiration from these works.
% the fast 2D object detector YOLO \cite{YOLO}. YOLO discretizes the camera image space into a 2D grid, where each grid cell is allowed to make a fixed number of 2D bounding box detections, from which only a few detections are selected, based on their confidence score and intersection-over-union (IoU) with other objects.
We divide the space around the camera into a 3D grid, that starts by subdividing the camera image into a 4x4 grid and extends this outwards into the 3rd (\emph{z}) axis. Each subdivision or \emph{cuboid} in our 3D grid is each allowed to predict N 3D object poses and their confidence scores. However, in contrast to \cite{YOLO,brazil2019m3d}, our network is anchor-free. Anchor-based approaches are limited to predicting objects whose sizes and aspect ratios resemble the anchor, and the number of objects per grid cell is limited to the number of anchors allowed for that cell. In contrast, each cuboid in our method is allowed to regress 3D bounding boxes of any size, shape, orientation and relative depth for N objects.

% Our work also resembles OFT-Net \cite{roddick2018orthographic}, which divides the 3D space around the camera into voxels and pools those 2D images features into the voxels. However, this is an intermediate representation and OFT-Net finally projects these features into an orthographic ground-plane grid, with each 2D (ground plane) grid cell being tasked with detecting the position, dimension, orientation and confidence of 3D bounding boxes. 

A major advantage of methods, including ours, that operate in the 3D voxel space \cite{qi2017pointnet,zhou2018voxelnet} or indeed the orthographic/BEV/ground-plane space \cite{wang2019pseudo,yang2018pixor,you2019pseudo,roddick2018orthographic}, over methods that operate in the camera image space \cite{mousavian20173d,qin2019monogrnet,chabot2017deep,he2019mono3d++} is that they can detect objects farther away from the camera, and we further extend this advantage and demonstrate that the accuracy of our method remains constant with distance from the camera.

% TODO: Add MonoDIS and ROI-10D.

% Single-stage detectors:
% M3D-RPN \cite{brazil2019m3d}.
% OFT-net \cite{roddick2018orthographic}.

% Multiple images:
% Joint-mono 3D detection and tracking \cite{hu2019joint}.

% Something about single shot detectors versus 2-stage detectors that include a Region Proposal Net. Seque into YOLO and how this is inspired by YOLO but uses pre-training for RGB-to-depth.
% Can add connection to OFT-net in the way 3D space is discretized.

\section{Method}

\begin{figure}[h]
\centering
    \centering
    \includegraphics[width=0.95\linewidth]{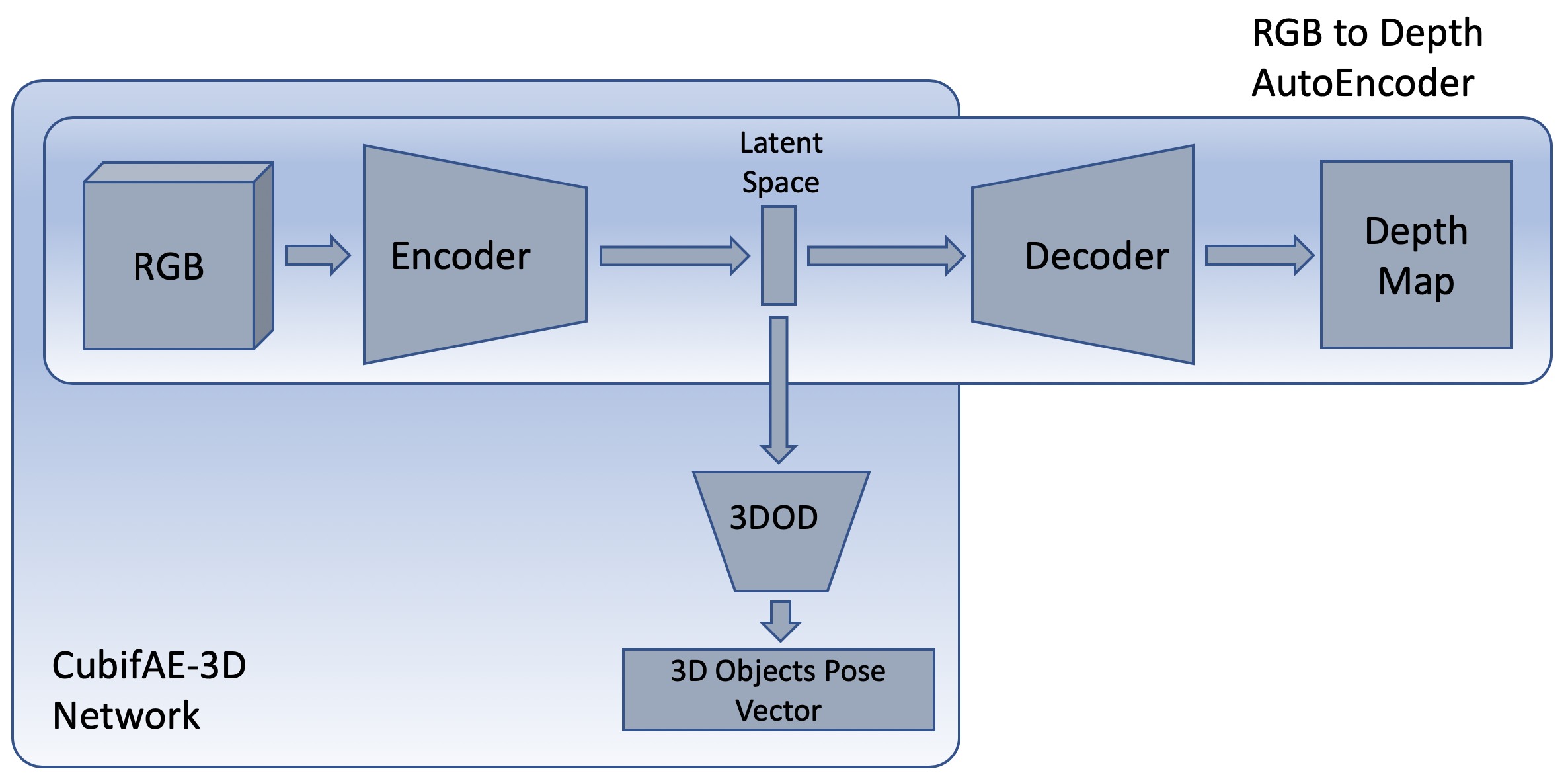}
\caption{CubifAE-3D high-level architecture
\label{fig:complete_model}}
\vspace{-0.55cm}
\end{figure}

\begin{figure*}[t!]
\centering
    \centering
    \includegraphics[width=0.9\linewidth]{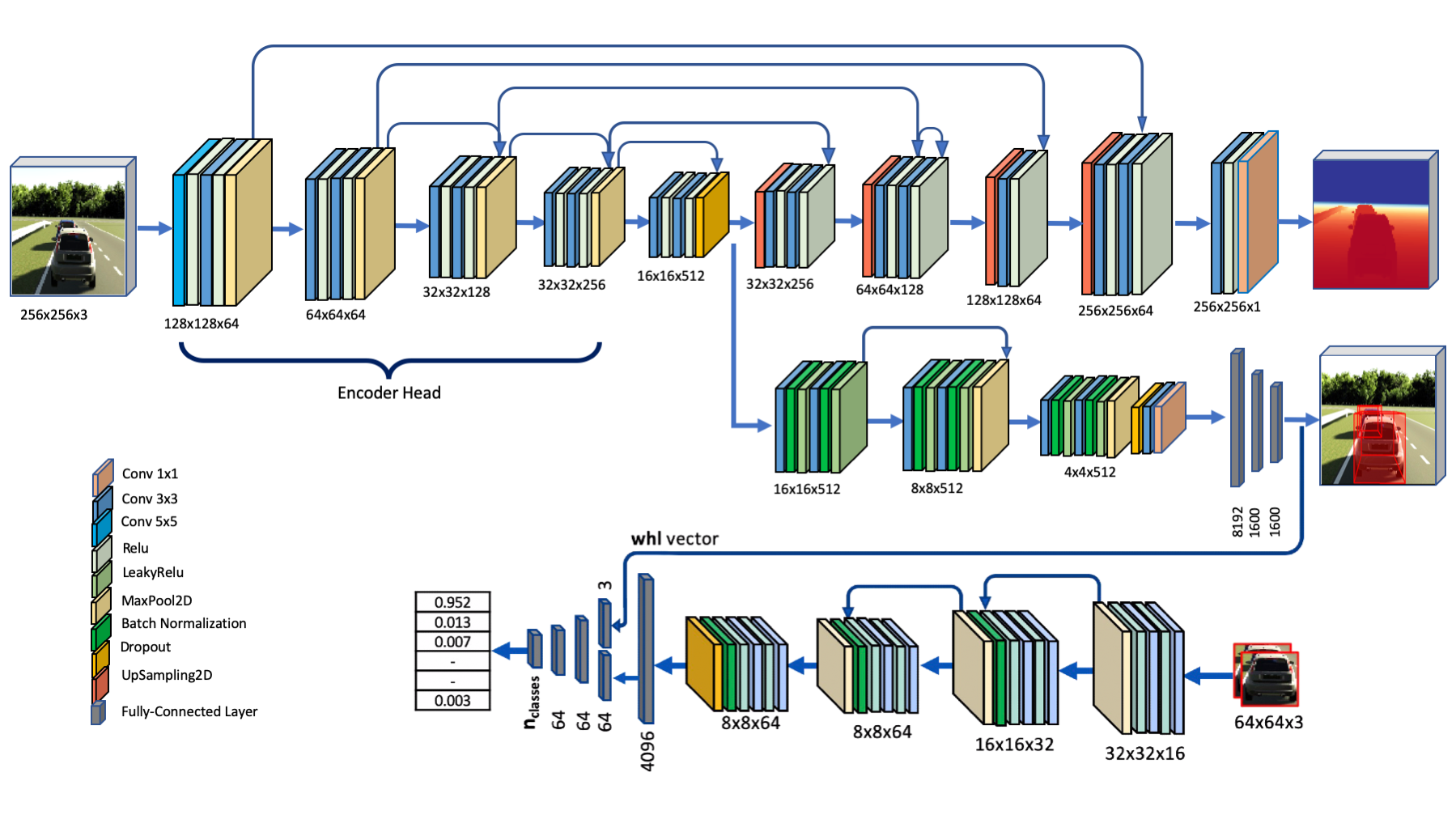}
\caption{Detailed model architecture of CubifAE-3D. The RGB-to-depth auto-encoder (top branch) is first trained in a supervised way with a combination of \emph{MSE} and \emph{Edge-Aware Smoothing Loss}. Once trained, the decoder is detached, encoder weights are frozen, and the encoder output is fed to the 3DOD model (middle branch), which is trained with a combination of \emph{$xyz_{loss}$}, \emph{$whl_{loss}$}, \emph{$orientation_{loss}$}, \emph{$iou_{loss}$}, and \emph{$conf_{loss}$}. A 2D bounding-box is obtained for each object by projecting its detected 3D bounding-box onto the camera image plane, cropped, and resized to 64x64 and fed to the \emph{classifier} model (bottom branch) along with the normalized \textbf{whl} vector for class prediction. The dimensions indicated correspond to the output tensor for each block. In our experiments, we also replace the \emph{encoder head} by a pretrained backbone network (VGG-16) and observe an improved performance.
\label{fig:cubifae-3d_architecture}}
\end{figure*}

Our method of performing 3D object detection relies on first learning the latent space embeddings for per-pixel RGB-to-depth predictions in an image. This is achieved by training an auto-encoder to predict the dense depth map from a single RGB image. Once trained, the decoder is detached, and the latent space embedding is fed to our 3DOD network. 
% A detailed model architecture diagram is shown in figure \ref{fig:cubifae-3d_architecture}, where the first row shows the rgb-to-depth auto-encoder.
The high-level model architecture is shown in figure \ref{fig:complete_model} and a detailed model architecture is shown in figure \ref{fig:cubifae-3d_architecture}. 

By training the auto-encoder first, we force its latent space to learn a compact RGB-to-depth embedding representation which is encoded in the latent space. A model which then operates on these encodings is thus able to formulate a relationship between the structures present in the RGB image and its real world depth. We then \emph{cubify} the monocular camera space and train our 3DOD model to detect object poses (\emph{$x_{center}$, $y_{center}$, $z_{center}$, $width$, $height$, $length$, $orientation$}). So, at the test time, only an RGB image is needed for detecting object poses. \emph{Orientation} term in the predicted pose refers to the rotation about \emph{y}-axis in the camera frame. For the sake of simplicity, we assume rotation about the other two axes to be \emph{zero}. An additional \emph{classifier} network with a small number of parameters is then used to classify all of these detected objects at once by resizing and stacking the object crops and feeding them to the \emph{classifier} model. This is done instead of predicting the classes directly as a part of the vector corresponding to each object from the 3DOD network in the favour of reducing number of parameters in the fully-connected layers and hence the inference time. We apply \emph{non-max suppression} to further filter out the object pose predictions with high \emph{IoU} by retaining only the objects with the highest \emph{confidence}.

In our experiments, we train the RGB-to-depth AE only in simulation and do not use/require dense depth information from the real datasets. 
% show that we only need the ground-truth dense depth map from simulation to train a RGB-to-depth auto-encoder. 
We realize that an RGB-to-depth model trained on one dataset is incompatible with another because of the camera intrinsics mismatch. However, because we only use the encoder part of the RGB-to-depth AE as an embedding, our 3DOD network automatically learns to compensate for this. We train the latent representation once, using one set of simulation data and use the same latent representation in the 3DOD network which then learns %the depth scale as we train it
to predict object poses on new real data. We also experiment with replacing the \textit{encoder head} by a pre-trained backbone model (VGG-16 \cite{simonyan2015deep}) and observe an improved performance over our baseline \textit{CubifAE-3D} model.  

\subsection{RGB-to-Depth Auto-Encoder}

The RGB-to-depth auto-encoder 
% as shown in figure  \ref{fig:complete_model} 
is trained with a combination of a Mean Squared Error (eqn \ref{eqn:mse_loss}) and an Edge-Aware Smoothing Loss (eqn \ref{eqn:eas_loss}) to perform depth map prediction from monocular RGB images. The network contains a U-Net \cite{ronneberger2015unet} like encoder-decoder architecture with skip connections and is shown in the first row of figure \ref{fig:cubifae-3d_architecture}. 

\begin{equation}
\begin{aligned}  
mse_{loss} = 
    \frac{\lambda_{mse}}{u*v} 
    \sum_{i=0,j=0}^{u,v} 
    {(d_{i,j}-\hat{d}_{i,j})}^2
\end{aligned}
\label{eqn:mse_loss}
\end{equation}

\begin{equation}
\begin{aligned}    
eas_{loss} = 
    \frac{\lambda_{eas}}{u*v} 
    \sum_{i=0,j=0}^{u,v} 
    \mid \partial _x \hat{d}_{i,j} \mid 
    \textbf{\textit{e}}^{-\mid \partial _x I_{i,j} \mid} + \\
    \mid \partial _y \hat{d}_{i,j} \mid 
    \textbf{\textit{e}}^{-\mid \partial _y I_{i,j} \mid}
\end{aligned}
\label{eqn:eas_loss}
\end{equation}

\begin{equation}
\begin{aligned}    
depth_{loss} = 
    mse_{loss} + eas_{loss}
\end{aligned}
\label{eqn:depth_loss}
\end{equation}

\begin{figure}[h]
\centering
    \centering
    \includegraphics[width=0.95\linewidth]{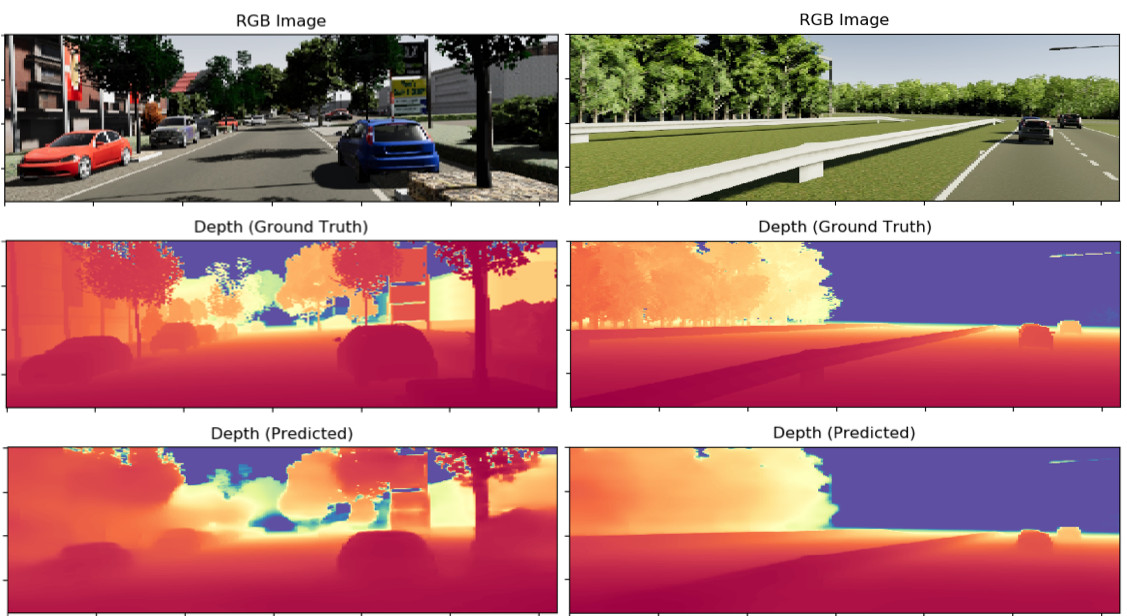}
\caption{Monocular RGB to Depth Map prediction for the vKITTI2 dataset. Top: input RGB images. Centre: ground-truth depth maps. Bottom: the depth map predictions from our RGB to Depth auto-encoder\label{fig:vkitti_rgb2depth}}
\vspace*{-0.5cm} 
\end{figure}

This model learns to predict accurate depth maps from monocular RGB images. The edge-aware smoothing loss penalizes high edge gradients if the image gradient is low and vice-versa, thus forcing the depth to be continuous and locally smooth within object boundaries while ensuring a clear depth-difference at the edges. This improves the accuracy of depth along the silhouette of the objects and avoids noisy holes within the object boundaries. Prediction results of this network is shown in figure \ref{fig:vkitti_rgb2depth}.

\subsection{The 3DOD Network}

Once the RGB-to-depth auto-encoder (AE) is trained, the head (encoder) of this network is detached, its weights are frozen, and the latent space is then fed to the 3DOD network for object pose estimation. This combined network 
% as shown in figure \ref{fig:3dod_model} 
thus learns to perform 3D object detection based on just a monocular RGB image as an input. We do not need the ground-truth depth map from a real dataset and instead pre-train the RGB-to-depth AE in simulation and then use the latent representation to train the 3DOD network. Ideally, the camera intrinsics between the simulated and real datasets need to match - as in our choice of datasets with \emph{KITTI} and its simulated counterpart, \emph{vKITTI2}. Experiments (not included in this paper) of using the pre-trained AE from vKITTI2 for 3DOD on a real dataset with a different camera, \emph{nuScenes} \cite{caesar2019nuscenes} resulted in reasonable results, but worse than SoTA. This means that given the ground truth 3D pose labels in the real dataset, the 3DOD network is able to compensate for the focal length mismatch to some extent, but there is a degradation of accuracy. We hope to further explore this in future work.%, but for now require that the simulated and real datasets have the same cameras.
% real datasets such as \emph{nuScenes} and \emph{KITTI}; we instead pre-train the RGB-to-depth AE in simulation and then use the latent representation to train the 3DOD network. The camera intrinsics between the simulated and real datasets do not match, but our 3DOD network, given ground truth 3D pose labels for the real dataset, automatically learns to compensate for the depth scale difference caused by the focal length mismatch.

% \begin{figure}[h]
% \centering
%     \centering
%     \includegraphics[width=0.65\linewidth]{fig/3dod_model.jpg}
% \caption{CubifAE-3D Model Architecture (encoder weights are frozen during the training of the 3DOD network).\label{fig:3dod_model}}
% \vspace*{-0.2cm} 
% \end{figure}

We prepare training labels for the 3DOD network in a way that allows each part of the network to 
%the parts of network 
only be responsible for detecting objects within a certain physical space relative to the ego-vehicle camera. We \emph{cubify} the 3D region-of-interest (ROI) of the ego-camera into a $3$-dimensional grid. This 3D grid is of size $4xM$, where the camera image plane is divided into $4$ regions along the (\emph{x,y}) dimensions of the camera coordinate frame, with \emph{z} axis further quantified into $M$ \emph{cuboids} for each of these $4$ regions.
% We do that by first dividing the entire 3D region-of-interest (ROI) into 4 grids/quads in the \emph{x} and \emph{y} directions where, each grid is further \emph{cubified} by dividing it into M cuboids in the z direction. 
Each cuboid in this $4xM$ grid is responsible for predicting up to $N$ objects in an increasing order of \emph{z} (depth) from the center of the ego-camera. The model predicts a vector of length $8$ for each possible object ($confidence$, $x_{center}$, $y_{center}$, $z_{center}$, $width$, $height$, $length$, $orientation$). This results in the model output being a vector of size $4$x$M$x$N$x$8$. The intuition behind dividing the visible ROI into a 4-dimensional grid in the \emph{x} and \emph{y} directions comes from the fact that the center of the image very closely corresponds to the optic center of the lens, and hence a 4x4 grid in the image space demarcates the 3D space in the same directions. This simplifies our normalization process given the \emph{x} and \emph{y} maximum limits. This also ensures that each grid in the camera plane contains an object within the corresponding 3D space; for example, the top-left grid contains objects with their center at \emph{negative x} and \emph{negative y} 3D coordinates, and the top-right grid contains objects with their center at \emph{positive x} and \emph{negative y} 3D coordinates. 

\begin{figure}[h]
\centering
    \centering
    \includegraphics[width=0.99\linewidth]{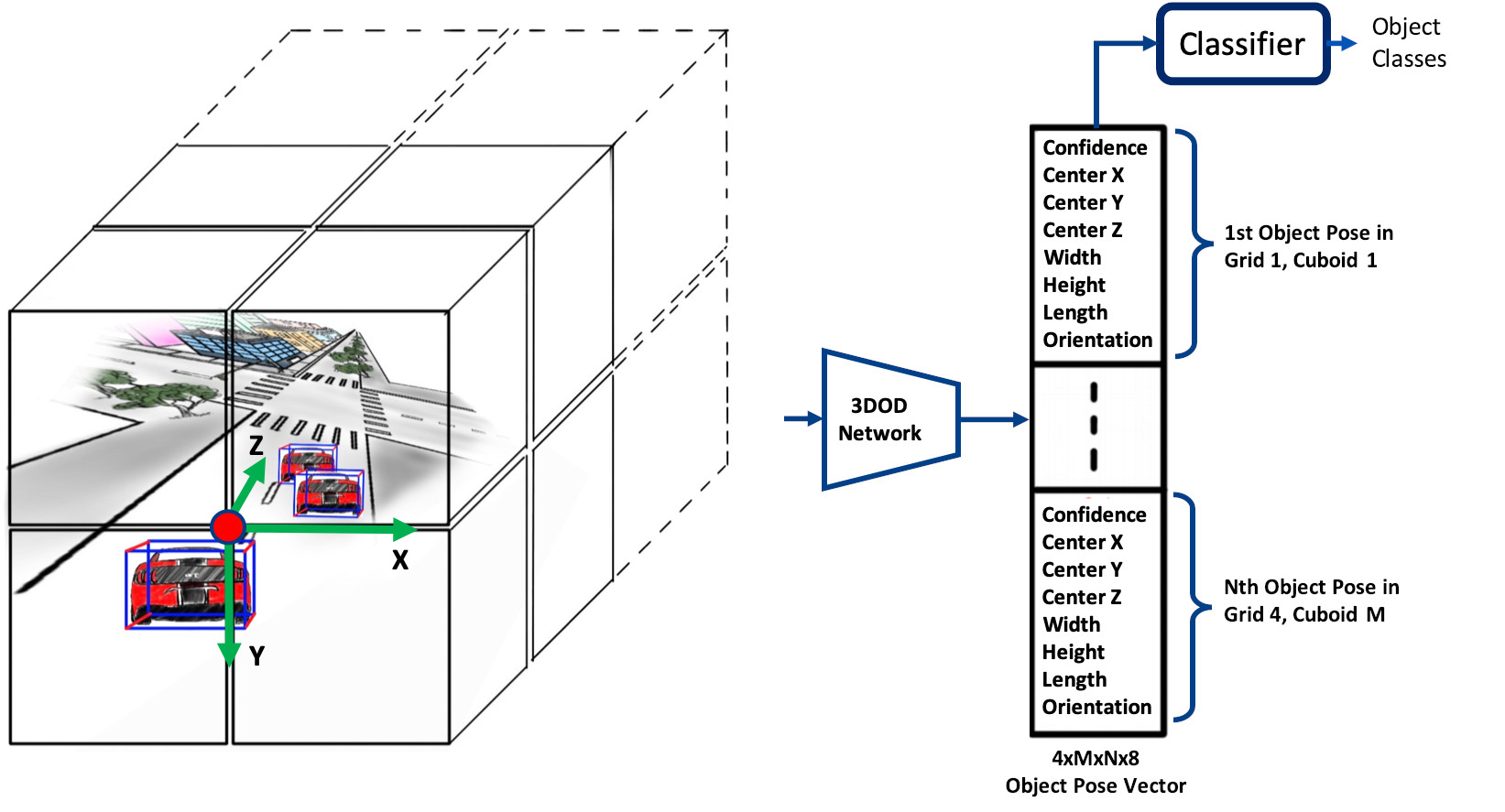}
\caption{\emph{Cubification} of the camera space: The perception region of interest is divided into a $4$x$4$x$M$ grid ($4$x$4$ in the \emph{x} and \emph{y} directions aligned with the camera image plane, where each grid has stacked on it, M cuboids in the \emph{z} direction). Each cuboid is responsible for predicting up to N object poses. The object coordinates and dimensions are then normalized between 0 and 1 in accordance with a prior that is computed from data statistics.\label{fig:camera_world}}
\vspace*{-0.3cm} 
\end{figure}

This \emph{Cubification} of the camera space is visualized in figure \ref{fig:camera_world}. We assume the standard camera coordinate system for our work (as used in computer vision applications), where, \emph{x} points to the right and is aligned with the image x-axis, \emph{y} points downward and is aligned with the image y-axis, and \emph{z} points in the direction that the camera is pointed at. 

We normalize 3D coordinates of the center of the object (\emph{x, y, z}) and dimensions (\emph{width, height, length}) between (\emph{$0$}, \emph{$1$}) in accordance with a prior that is computed from data statistics. Orientation is normalized by simply scaling the values from (\emph{$-\pi$}, \emph{$\pi$}) to (\emph{$0$}, \emph{$1$}). We use a Sigmoid activation function at the output of the final fully-connected layer for these predictions. Each predicted object ROI with high confidence is then projected to the 2D camera plane, cropped, and resized to a \emph{64x64} patch, which is then fed to a \emph{classifier} network that predicts a class for the detected object. 
% All object patches are further stacked and fed to a \emph{classifier} network which then predicts a class for each of those objects. 

The total loss function for this model is a weighted sum of 5 individual loss terms for minimizing the detection loss of object center coordinates, object dimensions, orientation, and their detection confidence. We take inspiration from YOLO \cite{YOLO} for the development of these loss functions. 

\begin{figure}[h]
\centering
    \centering
    \includegraphics[width=0.99\linewidth]{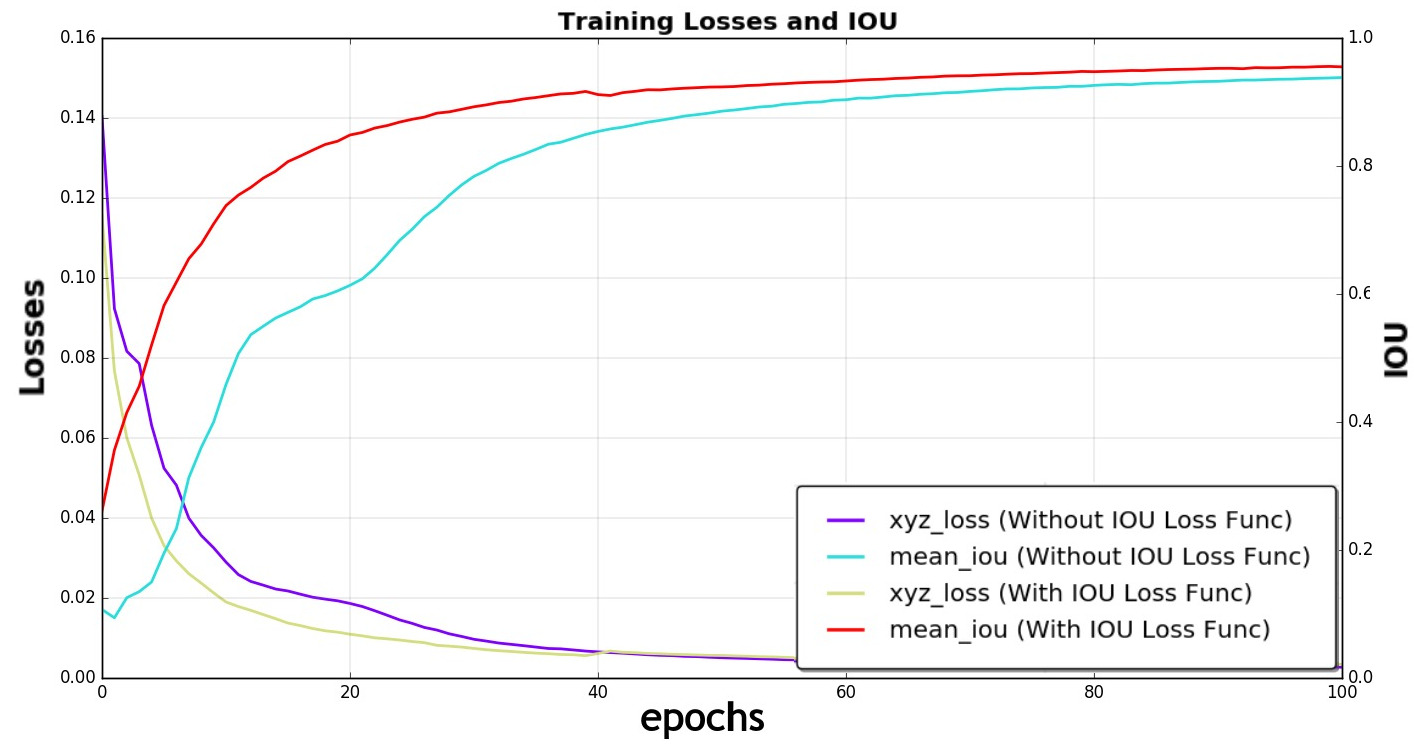}
\caption{Introducing the {$iou_{loss}$} improves error convergence and mean intersection-over-union (IoU). \emph{x}-axis shows number of training epochs, left \emph{y}-axis shows the \emph{$xyz_{loss}$} which is the mean-squared error of object center coordinate predictions (\emph{lower is better}), right \emph{y}-axis shows the mean \emph{3D intersection-over-union} between detected objects and ground-truth (\emph{higher is better}). \label{fig:iou_loss_comparision}}
\vspace*{-0.2cm} 
\end{figure}

Additionally, we also try to maximize 3D intersection-over-union (IoU) explicitly by minimizing the \emph{$iou_{loss}$} (eqn.\ref{eqn:iou_loss}) which is computed as negative log-likelihood of 3D IOU between ground-truth and prediction. Attempting to minimize this loss improves training by allowing the network to converge faster while improving the accuracy and mean IoU as shown in figure \ref{fig:iou_loss_comparision}. However, a very small weight, \emph{$\lambda_{iou}$} is chosen for this loss function to avoid the \emph{log} term from exploding.

\begin{equation}
\begin{aligned}    
xyz_{loss} = \\
    \frac{\lambda_{xyz}}{n_{true\_objs}}
    \sum_{i=0}^{4} \sum_{j=0}^{M} \sum_{k=0}^{N} 
    T_{ijk}^{n_{objs}}
    [{(x_{ijk} - \hat{x}_{ijk})}^2] + \\
    [{(y_{ijk} - \hat{y}_{ijk})}^2] + 
    [{(z_{ijk} - \hat{z}_{ijk})}^2]
\end{aligned}
\label{eqn:xyz_loss}
\end{equation}

\begin{equation}
\begin{aligned}    
whl_{loss} = \\
    \frac{\lambda_{whl}}{n_{true\_objs}}
    \sum_{i=0}^{4} \sum_{j=0}^{M} \sum_{k=0}^{N} 
    T_{ijk}^{n_{objs}}
    [{(\sqrt{w_{ijk}} - \sqrt{\hat{w}_{ijk}})}^2] + \\
    [{(\sqrt{h_{ijk}} - \sqrt{\hat{h}_{ijk}})}^2] + 
    [{(\sqrt{l_{ijk}} - \sqrt{\hat{l}_{ijk}})}^2]
\end{aligned}
\label{eqn:whl_loss}
\end{equation}

\begin{equation}
\begin{aligned}    
orientation_{loss} = \\
    \frac{\lambda_{orientation}}{n_{true\_objs}}
    \sum_{i=0}^{4} \sum_{j=0}^{M} \sum_{k=0}^{N} 
    T_{ijk}^{n_{objs}}
    [{(o_{ijk} - \hat{o}_{ijk})}^2]
\end{aligned}
\label{eqn:orientation_loss}
\end{equation}

\begin{equation}
\begin{aligned}    
iou_{loss} = \\
    \frac{\lambda_{iou}}{n_{true\_objs}}
    \sum_{i=0}^{4} \sum_{j=0}^{M} \sum_{k=0}^{N} 
    -T_{ijk}^{n_{objs}}
    [\log{({iou}_{ijk})}]
\end{aligned}
\label{eqn:iou_loss}
\end{equation}

\begin{equation}
\begin{aligned}    
conf_{loss} = \\
    \frac{\lambda_{conf}}{4*N*M}
    \sum_{i=0}^{4} \sum_{j=0}^{M} \sum_{k=0}^{N} 
    T_{ijk}^{n_{objs}}
    [{(c_{ijk} - \hat{c}_{ijk})}^2] + \\
    (1 - T_{ijk}^{n_{objs}})
    [{(c_{ijk} - \hat{c}_{ijk})}^2]
\end{aligned}
\label{eqn:conf_loss}
\end{equation}

\begin{equation}
\begin{aligned}    
total_{loss} = 
    xyz_{loss} + whl_{loss} + orientation_{loss} + \\ 
    iou_{loss} + conf_{loss}
\end{aligned}
\label{eqn:total_loss}
\end{equation}

In equations [\ref{eqn:xyz_loss} - \ref{eqn:total_loss}], $M$ is the number of \emph{cuboids} for each (of the 4) 2D image grid-sections, $N$ is the maximum number of possible objects per cuboid, and $n_{true\_objs}$ is the number of ground-truth objects. $T_{ijk}$ in these equations indicates whether or not an object appears at that position in the ground-truth label vector, its value being $1$ if it does, $0$ otherwise.

\subsection{Classifier}

The output of our 3DOD model is a vector of objects poses along with a \emph{confidence} value for their prediction. We compute 8 vertices for each bounding-box and project them onto the 2D camera plane. A 2D bounding-box is then obtained for each detected object, cropped, and resized to a \emph{64x64} patch. All object crops are then stacked and fed to the \emph{classifier} network which is responsible for predicting a class for each detected object using \emph{Softmax} activation function. Additionally, we also feed the normalized \textbf{whl} (\emph{width, height, length}) vector for each detected object to the network which is concatenated with the output of first fully-connected layer as shown in figure \ref{fig:cubifae-3d_architecture}. This further helps the network establish a relationship between object classes and their dimensions. Our \emph{classifier} model has a few residual blocks followed by fully-connected layers and contains only $430k$ parameters.

\section{Experimental Details}

We start with training the RGB-to-depth auto-encoder on the \emph{Virtual KITTI 2} dataset by considering depths up to $100\ meters$. Any depth value higher than that is set to $100\ meters$. The dense depth map is then normalized between [$-0.5$, $0.5$]. Once trained, the pre-trained encoder of RGB-to-depth auto-encoder is used to train our 3DOD model. 
We perform 3D Object Detection experiments on the \emph{Virtual KITTI 2} \cite{cabon2020virtual}, and \emph{KITTI} \cite{Geiger2012CVPR} datasets. For \emph{Virtual KITTI 2}, we use 18,943 training and 1,049 validation frames for training. 1000 samples were randomly chosen from the dataset and separated as a test set. \emph{KITTI} 3D object detection dataset contains $7481$ training samples and $7518$ testing samples (test split). Following \cite{shi2019pointrcnn}, we divide the training samples into \textit{train} split (3712 samples)  and \textit{val} split (3769  samples) for evaluation of our method. During the training of our \emph{classifier} model, we randomly perturb the center of object position (\emph{x, y, z}) by adding a small amount of noise in the range of [$-0.2$ meters, $0.2$ meters] to improve the robustness of classification. We also perform data augmentation by flipping the images in a left-right fashion and obtaining $x_{center}$ $\leftarrow$ $x_{center}$, $orientation$ $\leftarrow$ ($\pi$ - $orientation$); further augmentation is obtained by adjusting saturation and brightness by first converting the RGB images into HSV colorspace and then multiplying the \emph{saturation} ($S$) and \emph{value} ($V$) channel by a factor randomly chosen between [$0.5, 1.5$], and then converting it back to RGB. Other types of commonly used data augmentation techniques such as cropping, rotation, and offsetting the image plane, would need more sophisticated algorithms because of the requirement for the determination of new object poses as a result of image warping and will be considered as future work.

For our work presented here, we use 5 cuboids per 2D (\emph{xy} plane) grid section - chosen after ablation study on model performance with various combinations (see table  \ref{tab:ablation_kitti} for details), and allow the detection of a maximum of 10 objects per cuboid. As a result, our model is able to predict up to \emph{$4(grids)$}x\emph{$5(cuboids/grid)$}x\emph{$10(objects/cuboid)$}=\emph{$200$} objects within a 100 meter range. We specify a region-of-interest (ROI) and only consider objects within the ROI for our experiments. Furthermore, we obtain a \emph{prior} on object dimensions from the dataset statistics. The ROI around the ego-camera used in our work is: [\emph{$x = 40m, y = 10m, z = 100m$}]. We compute \emph{priors} for each dataset separately and normalize them between $[0, 1]$ for training. \emph{Priors} used for the datasets were: 

\textbf{Virtual KITTI 2}: [\emph{$width_{min}=1.13$, $width_{max}=3.02$, $height_{min}=1.22$, $height_{max}=4.20$, $length_{min}=2.22$, $length_{max}=16.44$}].

% \textbf{nuScenes}: [\emph{$width_{min}=0.27$, $width_{max}=3.53$, $height_{min}=0.40$, $height_{max}=4.46$, $length_{min}=0.30$, $length_{max}=14.46$}].

\textbf{KITTI}: [\emph{$width_{min}=0.30$, $width_{max}=3.01$, $height_{min}=0.76$, $height_{max}=4.20$, $length_{min}=0.20$, $length_{max}=35.24$}].

Additionally, we also experiment by replacing the \emph{encoder head} with a pre-trained backbone network (VGG-16). This results in slightly improved performance as tabulated in table \ref{tab:quant_results_kitti}. For our VGG-16 based model, we do not use U-Net like skip connections between encoder and decoder; we extract all the VGG-16 layers up until the last $16$x$16$ spatial resolution layer (namely \emph{$block5\_conv3$}) to replace the \emph{encoder head} with. The choice of using 5 cuboids per grid and 10 objects per cuboid was made after an exhaustive search through various combinations of cuboids and objects per cuboid, shown in the ablation study table \ref{tab:ablation_kitti}. The table shows results for the \emph{car}, \emph{pedestrian} and \emph{cyclist} classes for 3D and BEV IoUs in the KITTI object detection eval kit. IoU thresholds for these classes for computing metrics were 
$0.7$, $0.5$ and $0.5$ respectively. 5 cuboids and 10 objects gave the best results for the car (easy), pedestrian (easy) and cyclist (easy, moderate, and hard) classes and difficulty levels. While 6 cuboids, 10 objects per cuboid gave better results on some classes, we chose the former to save on compute.
% Our ablation studies for choosing these parameters are summarized in table \ref{tab:ablation_kitti}.

% Cubification ablation study: results with varying \emph{cuboids} and \emph{objects} per grid. Green and blue highlights show the best performing variant for the BEV and 3D IOUs in the KITTI object detection dataset.

    %   Ablation study of CubifAE-3D \textit{(w/vgg-16)} on the KITTI \emph{validation} split for \emph{car}, \emph{pedestrian}, and \emph{cyclist}. These metrics are evaluated by $AP|_{R11}$. \textbf{$n_\emph{cuboids}$} refers to number of \emph{cuboids} per \emph{grid}, and \textbf{$n_\emph{objects}$} refers to the maximum number of objects that can be detected within each \emph{cuboid}. Maximum number of objects that can be detected by the model in each case is fiven by $4$x\textbf{$n_\emph{cuboids}$}x\textbf{$n_\emph{objects}$}. 3D and BEV signifies the type of IOU (3D vs. birds-eye-view) used by the KITTI object detection eval kit. IOU used for computing these these metrics are $0.7$, $0.5$, $0.5$ for \emph{car}, \emph{pedestrian}, and \emph{cyclist} respectively.

We train our models for $1500$ epochs with a batch size of 16, \emph{learning rate} of $10^{-4}$ and \emph{decay rate} of $0.001$. 
The \emph{Adam Optimizer} with $\beta_1=0.9$, $\beta_2=0.999$ is used for error optimization. \emph{Lambda} weights used for loss functions are: $\lambda_{mse}=0.8, \lambda_{eas}=0.2, \lambda_{xyz}=5.0, \lambda_{whl}=5.0, \lambda_{orientation}=1.0, \lambda_{iou}=0.01, \lambda_{conf}=0.5$. Our complete model (\emph{Encoder of RGB-to-depth AE + 3DOD + Classifier}), has $38.9M$ parameters and runs at $7.2$ FPS on a single NVIDIA RTX 2070 GPU during inference. To avoid overfitting, we use \emph{dropout} layers with dropout rate of $0.5$ in our auto-encoder as well as 3DOD model. In the \emph{classifier} model we employ \emph{L2 regularization} with a factor of $10^{-4}$. 

\section{Results}

\begin{figure}[!htb]
\centering
    \centering
    \includegraphics[width=0.35\linewidth]{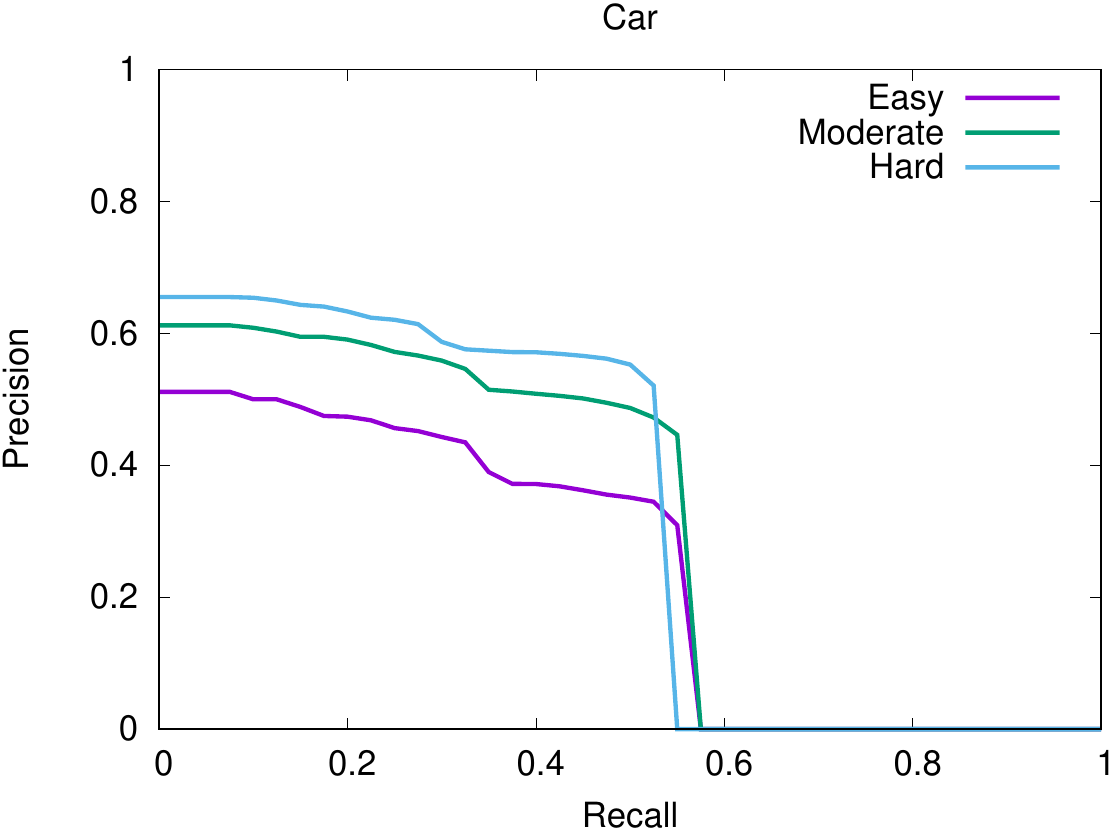}
    \hspace{-0.15in}
    \includegraphics[width=0.35\linewidth]{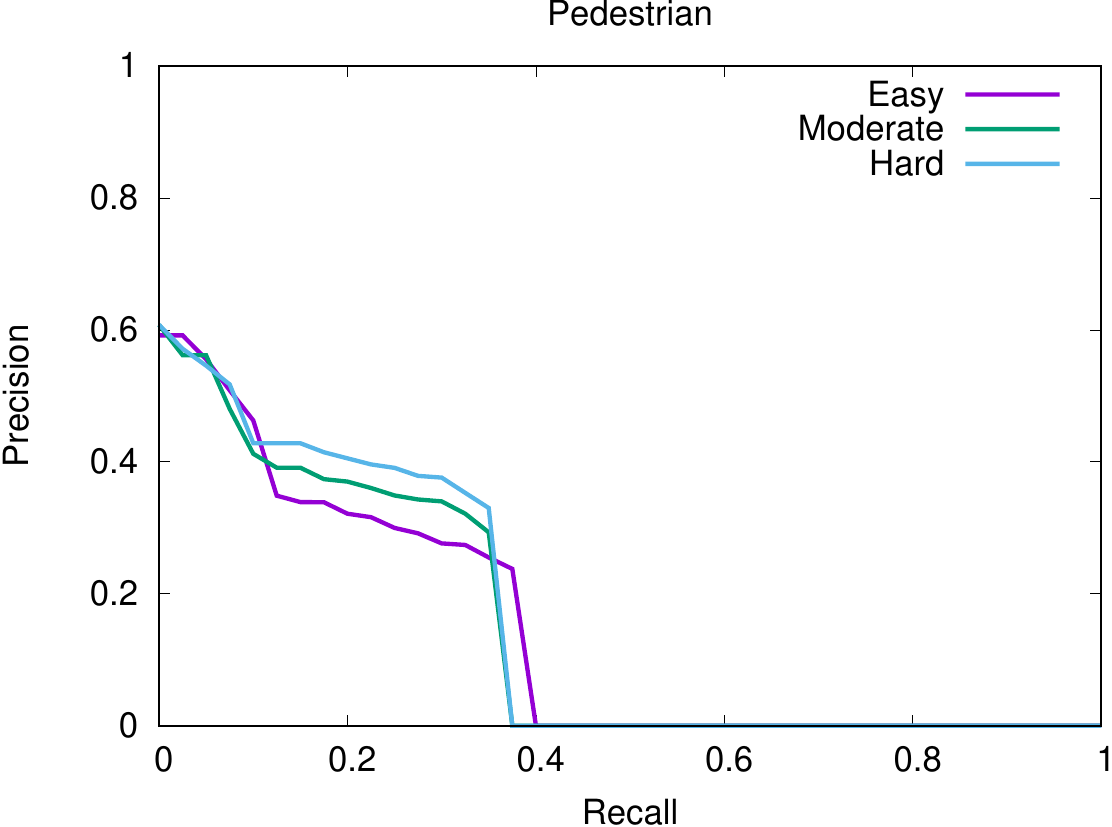}
    \hspace{-0.15in}
    \includegraphics[width=0.35\linewidth]{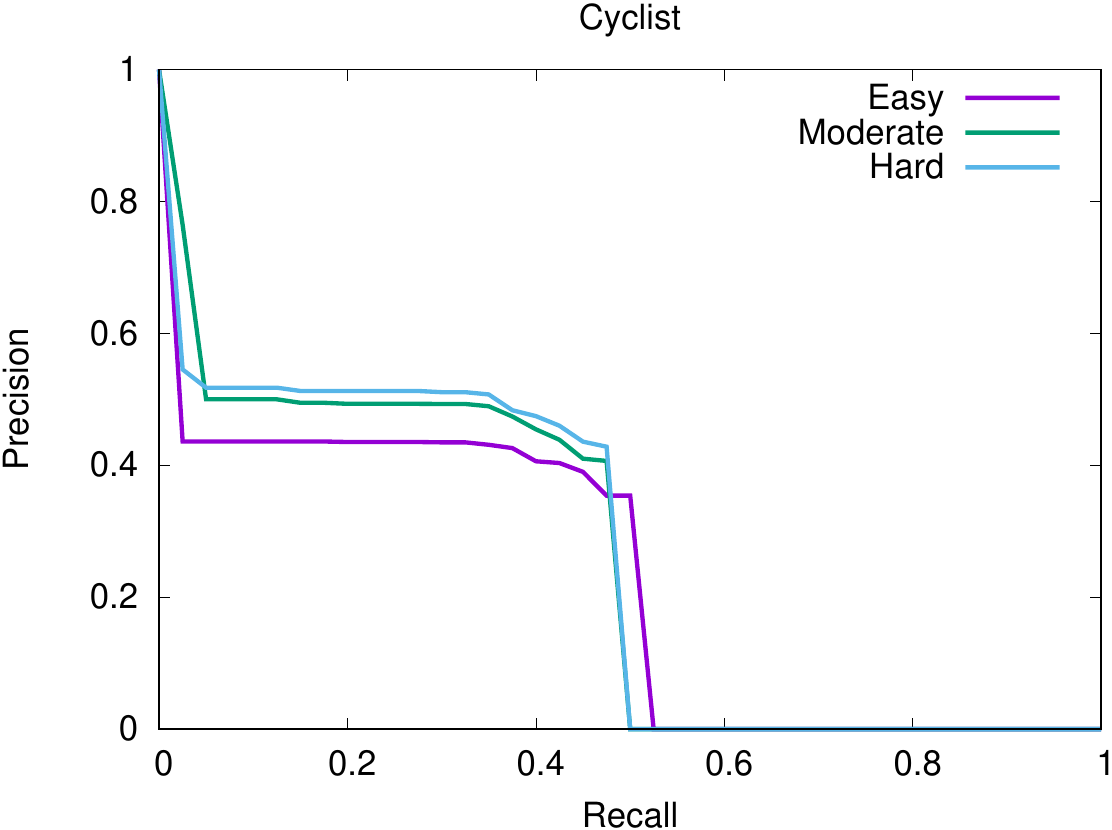}
\caption{Precision-Recall curves obtained using KITTI 3D object detection eval kit. These were computed on KITTI \textit{val} split for \textit{car}, \textit{pedestrian}, and \textit{cyclist} classes using 3D IOU of $0.7$, $0.5$, and $0.5$ respectively.} \label{fig:precision_recall_curve}
\end{figure}

\begin{figure*}[!h]
\centering
    \begin{subfigure}
    \centering
    \includegraphics[width=0.24\linewidth]{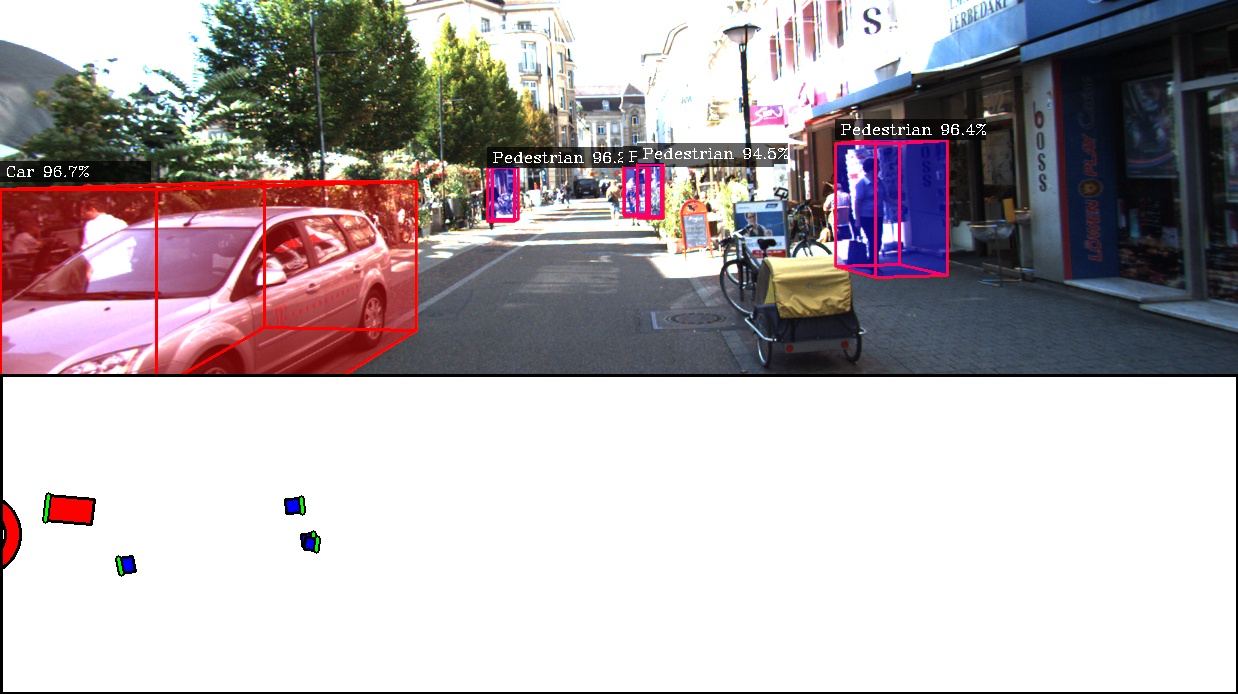}
    \end{subfigure}
    \begin{subfigure}
    \centering
    \includegraphics[width=0.24\linewidth]{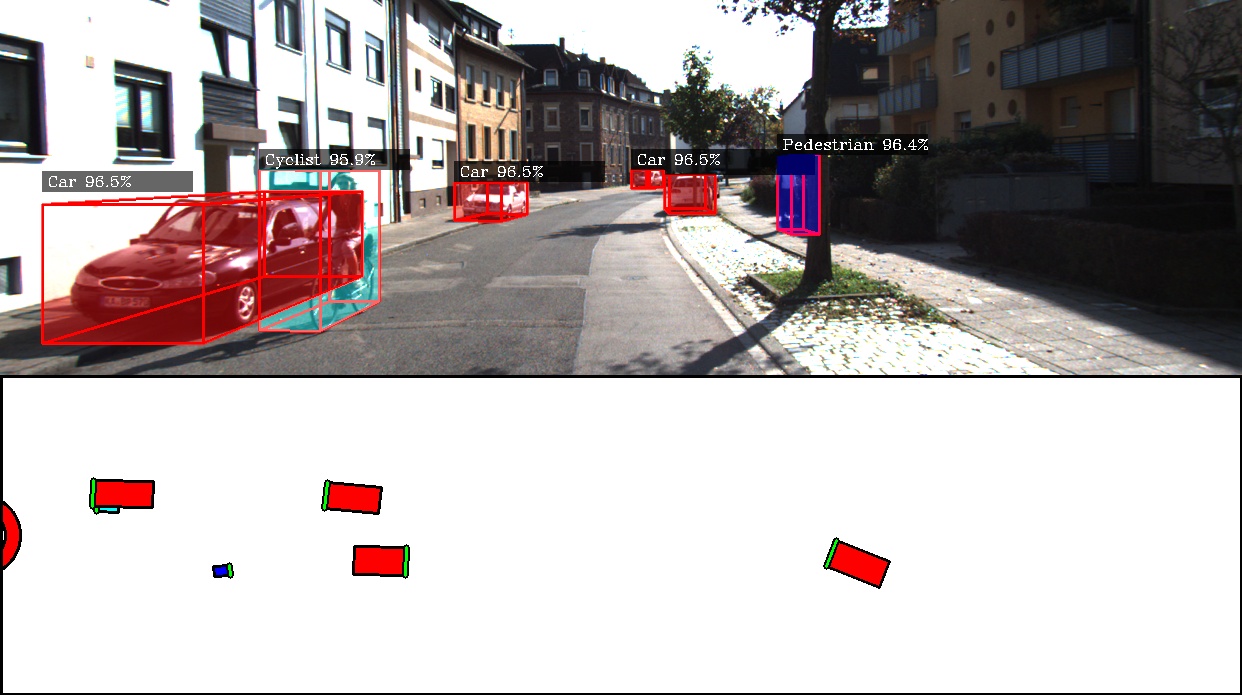}
    \end{subfigure}
    \begin{subfigure}
    \centering
    \includegraphics[width=0.24\linewidth]{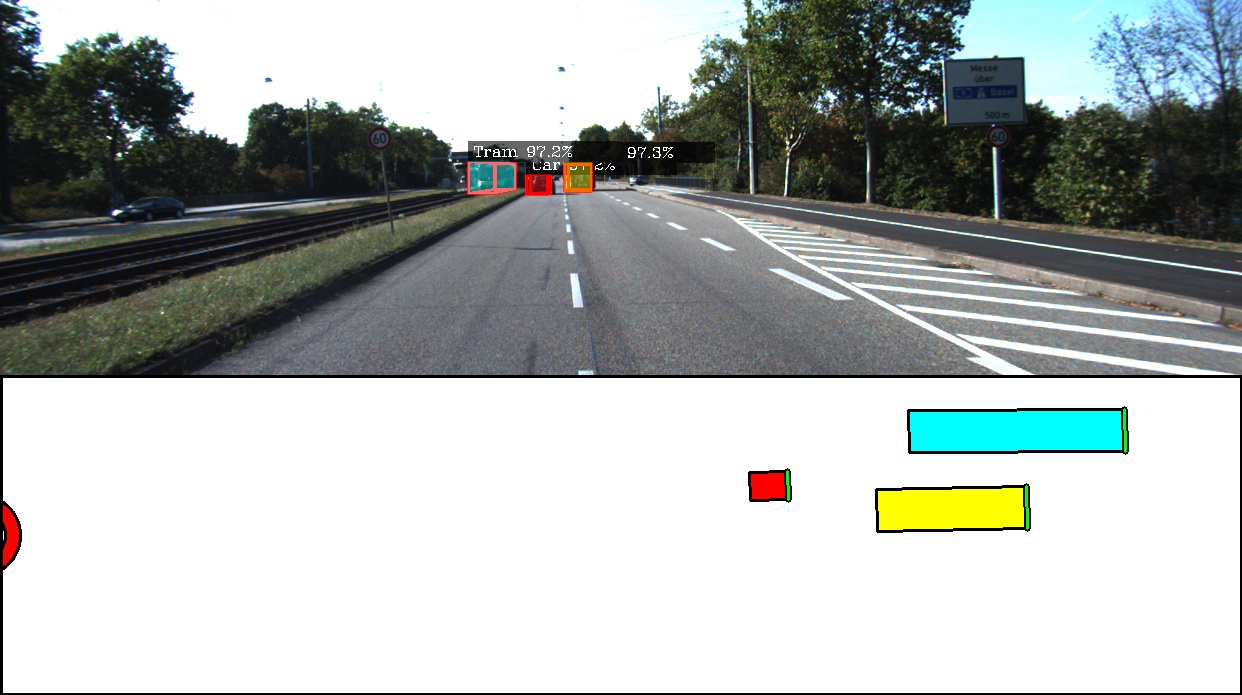}
    \end{subfigure}
    \begin{subfigure}
    \centering
    \includegraphics[width=0.24\linewidth]{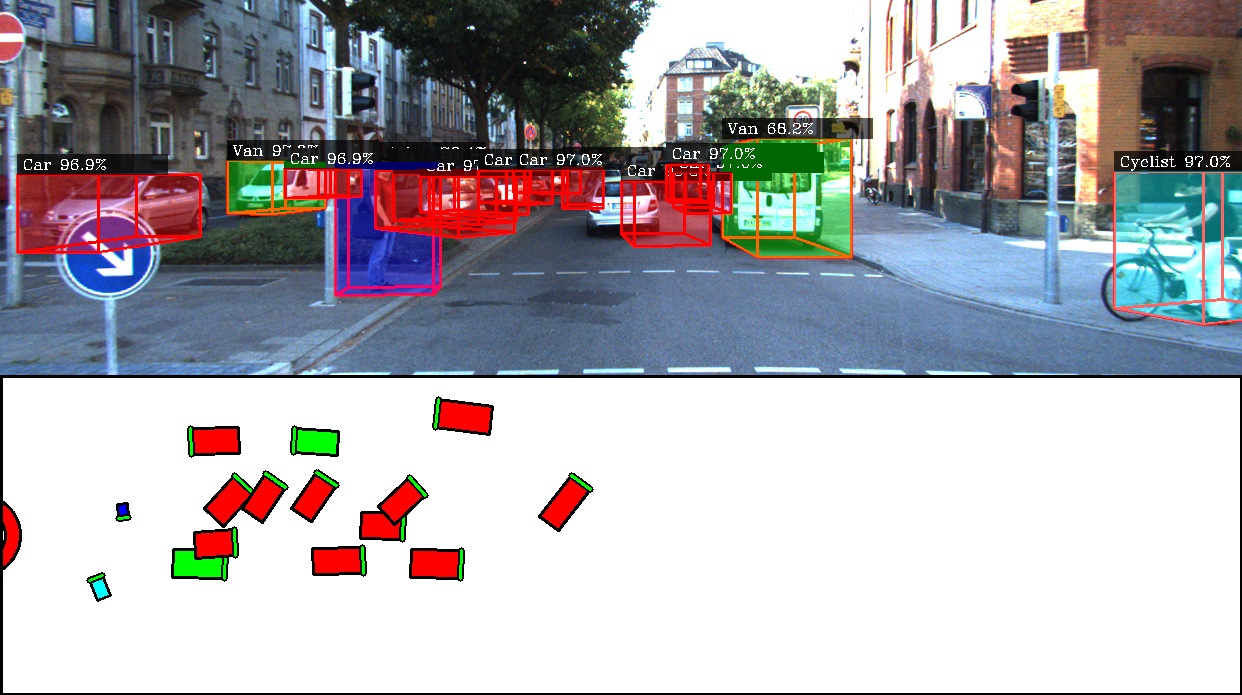}
    \end{subfigure}
    \begin{subfigure}
    \centering
    \includegraphics[width=0.24\linewidth]{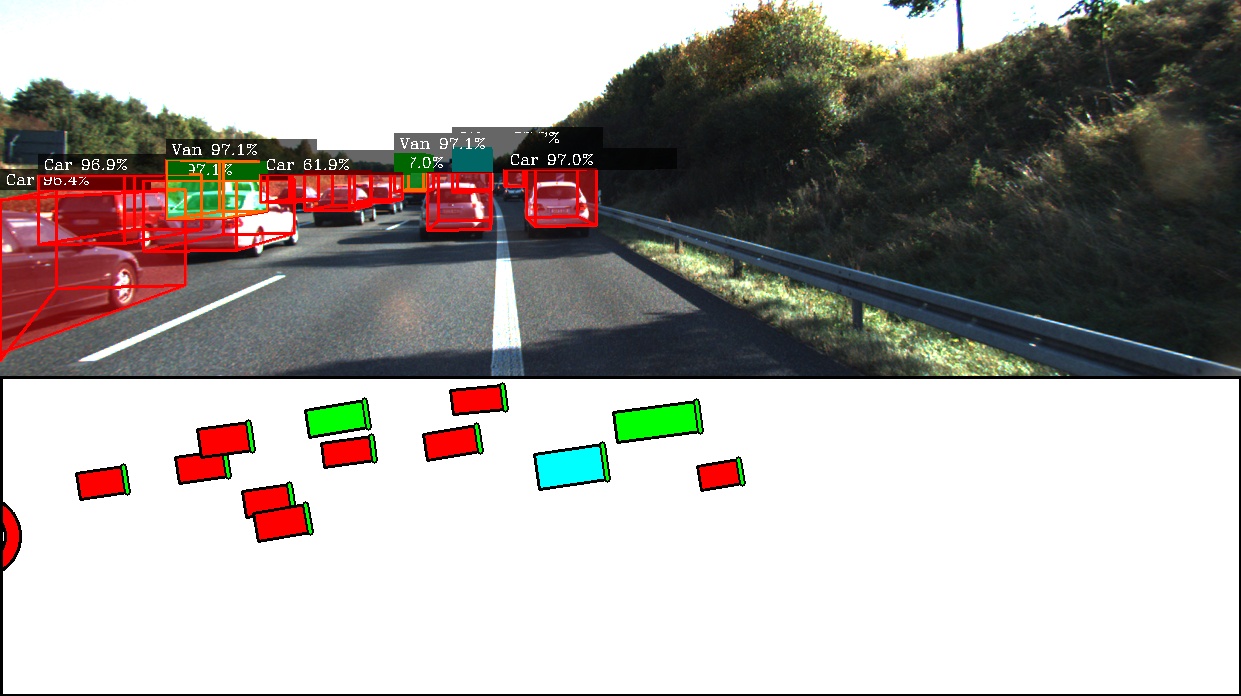}
    \end{subfigure}
    \begin{subfigure}
    \centering
    \includegraphics[width=0.24\linewidth]{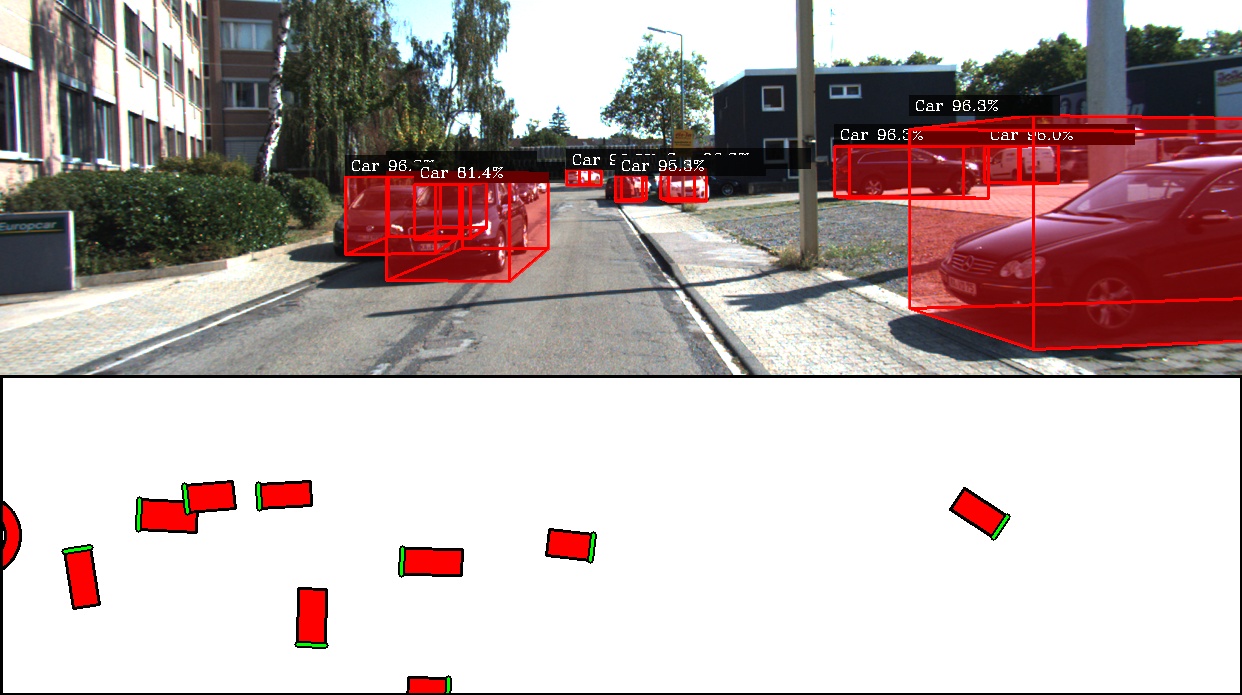}
    \end{subfigure}
    \begin{subfigure}
    \centering
    \includegraphics[width=0.24\linewidth]{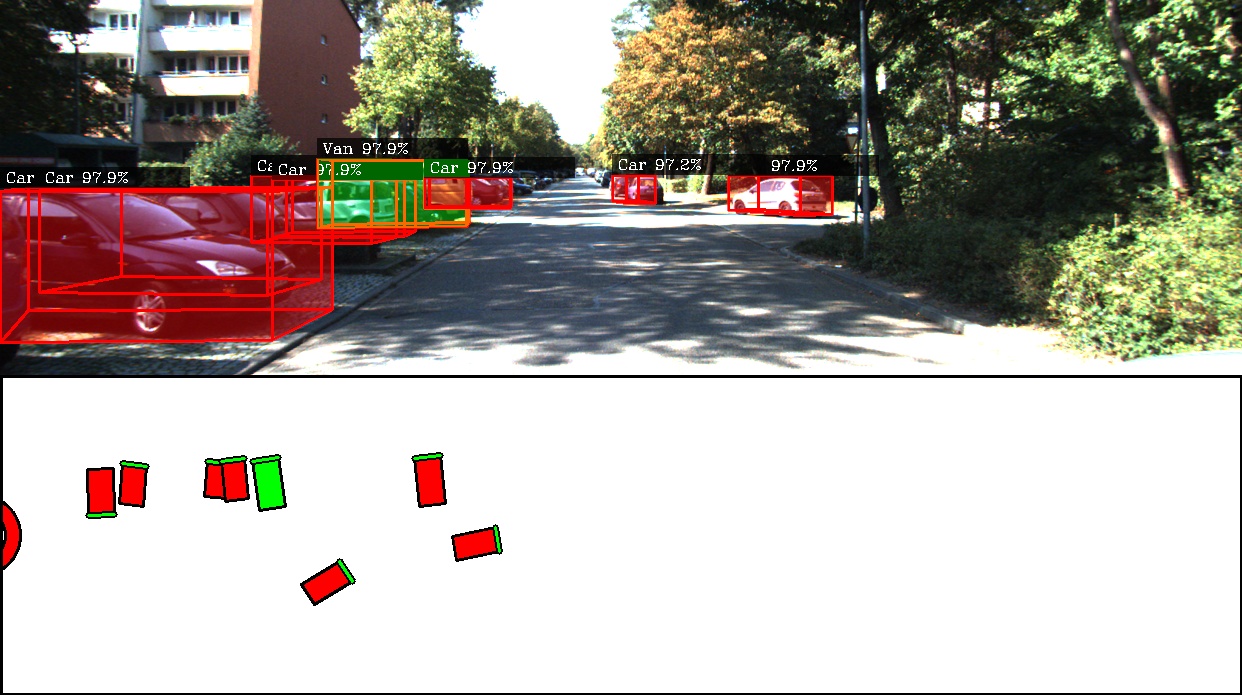}
    \end{subfigure}
    \begin{subfigure}
    \centering
    \includegraphics[width=0.24\linewidth]{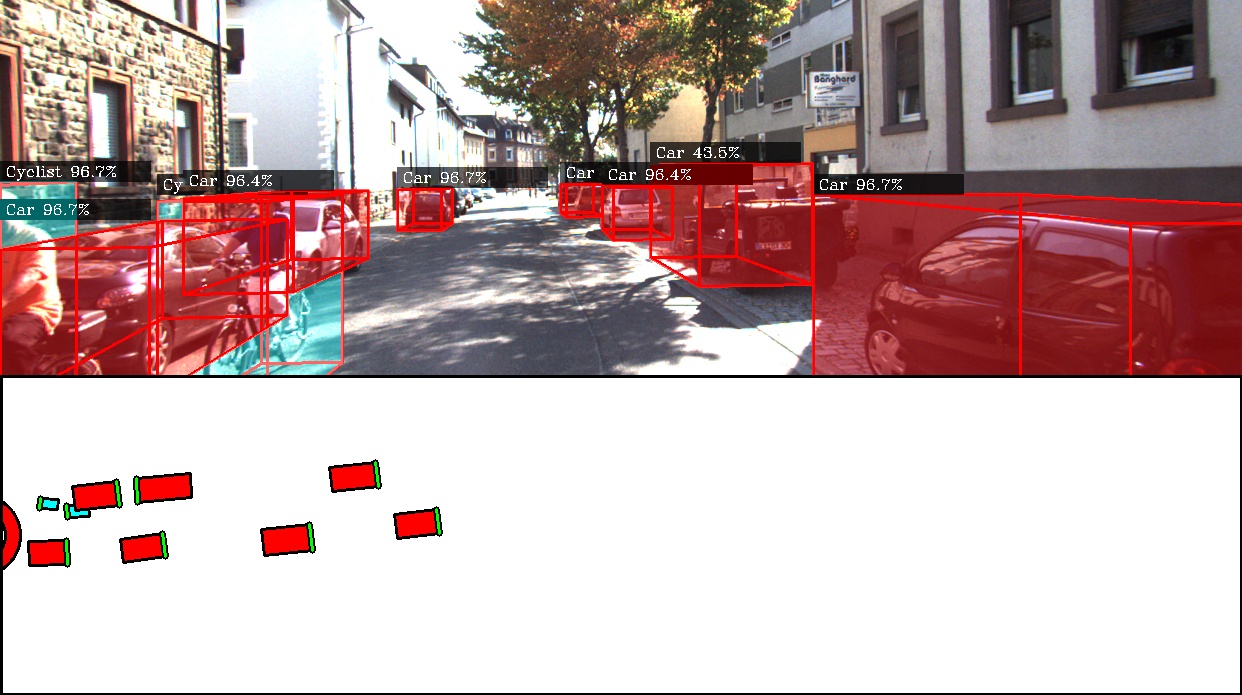}
    \end{subfigure}
\caption{Samples of qualitative results on the KITTI dataset. The top part of each image shows a bounding box obtained as a 2D projection of their 3D poses (\emph{red: car, green: van, blue: pedestrian, yellow: truck, cyan: cyclist, tram, others}). The bottom part shows a birds-eye view of the object poses with the ego-vehicle positioned at the center of red circle drawn on the left; pointing towards the right of the image. Front of each object is shown by a green line to indicate its orientation.
\label{fig:qual_res_kitti_main}}
\vspace*{-0.35cm}
\end{figure*}

We compute the \emph{Mean Average Precision} for the \emph{vKITTI2} and \emph{KITTI} datasets using 101-point interpolation as used in MS COCO \cite{lin2014microsoft} for \emph{3D IoU} thresholds of $0.3$, $0.5$, and $0.7$ by varying the \emph{confidence} threshold between [$0.0, 1.0$] across all classes as shown in table \ref{tab:quant_results} (\emph{number of classes: KITTI - 8, Virtual KITTI - 3}). Figure \ref{fig:precision_recall_curve} shows the Precision-Recall curves as computed using the KITTI object detection benchmark eval kit for 3D object detection problem (easy, medium, and hard difficulty levels). IOU thresholds used for obtaining these graphs were $0.7$, $0.5$, and $0.5$ for \emph{car}, \emph{pedestrian}, and \emph{cyclist} respectively. Additionally, we do comprehensive quantitative benchmarking and comparisions with other SoTA methods for the KITTI dataset and summarize them in table \ref{tab:quant_results_kitti}. Aside from the most recent M3D-RPN \cite{brazil2019m3d} work, we notice that our best-performing method (with 5 cuboids, 10 objects per cuboid and the vgg-16 head), is able to compete with other SoTA techniques, especially in the moderate and hard difficulty levels, even beating M3D-RPN in the cyclist class. Our method is second-best in terms of moderate and hard difficulty levels for the car class (the only class reported in all competing methods except for M3D-RPN). 
% pretty well for objects with higher level of difficulties (represented by small number of pixels, higher occlusion) which has never been seen with other state-of-the-art methods.
Qualitative results of our model (\emph{w/vgg-16}) on the KITTI object detection dataset (\emph{val} split) is shown in figure \ref{fig:qual_res_kitti_main}. 

\begin{figure}[!tb]
\centering
    \centering
    \includegraphics[width=0.90\linewidth]{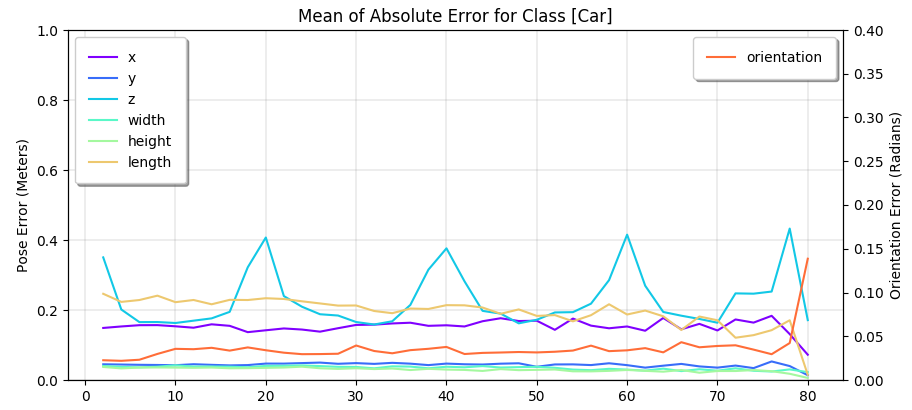}
\caption{Distribution of errors for the \emph{car} class. Because of the \emph{cubification} method, our errors are localized within each cuboid and does not increase with respect to absolute distance from the ego-vehicle. \label{fig:err_map_z}}
\vspace*{-0.55cm}
\end{figure}

\begin{table}[!htb]
    \begin{center}
      \caption{Monocular 3D Object Detection results on the vKITTI2 and KITTI datasets. $mAP_x$ stands for $\%$ of \emph{Mean Average Precision (mAP)} for \emph{3D IoU} threshold of \emph{x}. Higher is better.}
      \label{tab:quant_results}
      \begin{tabular}{l|l||l|l|l}
      \toprule
        \textbf{Method}&\textbf{Dataset}&$mAP_{\textbf{.3}}$&$mAP_{{\textbf{.5}}}$&$mAP_{\textbf{.7}}$ \\
        \midrule
        CubifAE-3D    &   vKITTI 2    &   86.6    &   66.7   &   34.1 \\
        %CubifAE-3D    &   nuScenes    &   77.6&   68.6   &  54.0 \\
        CubifAE-3D    &   KITTI      &   83.8    &   59.2   &   27.0 \\
        \bottomrule
      \end{tabular}
  \end{center}
  \vspace*{-0.55cm}
\end{table}

\begin{table}[!htb]
    \begin{center}
      \caption{Monocular 3D Object Detection Results on the KITTI \emph{validation} split for \emph{car}, \emph{pedestrian}, and \emph{cyclist}. These metrics are evaluated by $AP|_{R11}$ at {0.7} IoU threshold for \emph{car}, and {0.5} IoU threshold for \emph{pedestrian} and \emph{cyclist}. Higher is better. Best performing method is highlighted in bold and second best in green.}
      
      \label{tab:quant_results_kitti}
      \begin{tabular}{l|l||l|l|l}
      \toprule
        \textbf{Method}&\textbf{Class}&\textbf{Easy}&\textbf{Mod.}&\textbf{Hard} \\
        \midrule
        OFT-Net\cite{roddick2018orthographic} & Car    &   4.07   &   3.27    &   3.29    \\
        FQNet\cite{liu2019deep}   & Car   &   5.98    &   5.50    &   4.75    \\
        Mono3D++\cite{he2019mono3d++} & Car    &   10.60  &   7.90    &   5.70   \\
        % ROI-10D\cite{manhardt2018roi10d} &   Car  &  10.12    &  1.76    &   1.30    \\
        % \textit{(w/o depth)} & & & & \\
        % ROI-10D\cite{manhardt2018roi10d} &   Car  &  7.79    &   5.16    &   3.95    \\
        % \textit{(w/depth)} & & & & \\
        ROI-10D\cite{manhardt2018roi10d}  &   Car  &  9.61 &   6.63    &   6.29    \\
        % \textit{(w/depth, syn.)} & & & & \\
        MF3D\cite{Xu_2018_CVPR} & Car    &   10.53  &   5.69    &   5.39   \\
        GS3D\cite{Li_2019_CVPR} & Car    &   11.63  &   10.51    &   10.51   \\
        MonoGRNet\cite{qin2019monogrnet} & Car    &   13.88  &   10.19    &   7.62   \\
        Barabanau et al.\cite{barabanau2019monocular} & Car  &   \textcolor{PineGreen}{13.96}  &  7.37  &  4.54   \\
        M3D-RPN\cite{brazil2019m3d} & Car & \textbf{20.27} & \textbf{17.06} & \textbf{15.21} \\
        M3D-RPN\cite{brazil2019m3d} & Ped. & - & \textbf{11.28} & - \\
        M3D-RPN\cite{brazil2019m3d} & Cyclist & - & \textcolor{PineGreen}{10.01} & - \\
        \midrule
        ours  &   Car    &   7.94  &  9.21  &    10.43 \\
        ours  &   Ped.    &   4.64  &   4.79  &   4.96 \\
        ours  &   Cyclist    &   8.27 &   8.61  &   9.01 \\
        \midrule
        ours \textit{(w/vgg-16)} &   Car    &   9.89  &    \textcolor{PineGreen}{11.20}  &    \textcolor{PineGreen}{12.33} \\
        ours \textit{(w/vgg-16)} &   Ped.    &   6.74  & 6.82  &   7.04 \\
        ours \textit{(w/vgg-16)} &   Cyclist    &   12.93  &    \textbf{12.18}  &   12.09 \\
        \bottomrule
      \end{tabular}

  \end{center}
  \vspace*{-0.55cm}
\end{table}

\begin{table*}[!t]
    \begin{center}
      \caption{Cubification ablation study: results with varying \emph{cuboids} and \emph{objects} per grid. Green and blue highlights show the best performing variant for the BEV and 3D IOUs respectively in the KITTI object detection dataset.}

      \label{tab:ablation_kitti}
      \begin{tabular}{l|l|l||l|l|l}
      \toprule
        \textbf{$n_\emph{cuboids}$} & \textbf{$n_\emph{objects}$} & 3D / BEV & Car (Easy/Mod./Hard) & Pedestrian (Easy/Mod./Hard) & Cyclist (Easy/Mod./Hard) \\
        \midrule
        4 & 10 & 3D  &   $5.30$ / $7.40$ / $8.32$ & $2.56$ / $2.47$ / $2.78$ & $4.91$ / $5.73$ / $5.91$ \\
        4 & 10 & BEV &   $7.40$ / $9.97$ / $10.95$ & $2.74$ / $2.79$ / $3.02$ & $5.74$ / $6.85$ / $7.27$ \\
        4 & 12 & 3D &   $4.69$ / $7.11$ / $7.01$ & $0.098$ / $1.38$ / $1.31$ & $2.28$ / $2.97$ / $2.75$ \\
        4 & 12 & BEV &   $6.59$ / $9.77$ / $10.41$ & $1.39$ / $1.92$ / $2.02$ & $3.27$ / $4.46$ / $4.48$ \\
        6 & 10 & 3D  &   $8.97$ / \textcolor{Blue}{$11.44$} / \textcolor{Blue}{$12.47$} & $6.63$ / \textcolor{Blue}{$7.31$} / \textcolor{Blue}{$7.29$} & $7.60$ / $9.20$ / $9.45$ \\
        6 & 10 & BEV &   \textcolor{PineGreen}{$11.40$} / \textcolor{PineGreen}{$14.09$} / \textcolor{PineGreen}{$15.18$} & $7.66$ / $8.05$ / \textcolor{PineGreen}{$8.74$} & $8.70$ / $10.88$ / $11.24$ \\
        6 & 8 & 3D &   $4.57$ / $5.73$ / $6.33$ & $3.05$ / $2.76$ /  $2.61$ & $5.25$ / $4.49$ / $4.73$ \\
        6 & 8 & BEV &   $7.41$ / $9.63$ / $10.56$ & $3.99$ / $3.86$ / $3.57$ & $6.56$ / $6.06$ / $6.52$ \\
        5 & 10 & 3D &   \textcolor{Blue}{$9.89$} / $11.20$ / $12.33$ & \textcolor{Blue}{$6.74$} / $6.82$ / $7.04$ & \textcolor{Blue}{$12.93$} / \textcolor{Blue}{$12.18$} / \textcolor{Blue}{$12.09$} \\
        5 & 10 & BEV &   $11.12$ / $13.82$ / $14.95$ & \textcolor{PineGreen}{$7.95$} / \textcolor{PineGreen}{$8.14$} / $8.21$ &  \textcolor{PineGreen}{$14.07$} /  \textcolor{PineGreen}{$14.18$} /  \textcolor{PineGreen}{$13.77$} \\
        \bottomrule
      \end{tabular}

  \end{center}
  \vspace*{-0.1cm}
\end{table*}

We also attempt to model the error distribution with respect to the absolute distance from the ego-vehicle for each class. We find that, because of our \emph{cubification} techniques, the error is highly localized within each \emph{cuboid} and does not increase with increasing distance from the ego-vehicle. Errors are found to be the minimum at the center of each \emph{cuboid} and maximum at the beginning and end of each \emph{cuboid} for \emph{cars} as evident from figure \ref{fig:err_map_z}. This figure shows the plot of error distribution for the \emph{car} and \emph{pedestrian} class. Note that the \emph{cuboid} size in the depth ($z$) direction was chosen to be $20\ meters$ and hence error peaks in \emph{z} can be seen at $0, 20, 40, 60, 80, and\ 100\ meters$.

% We show qualitative results of our model on \emph{KITTI} dataset in figure \ref{fig:qual_res_kitti}, \emph{nuScenes} dataset in figure \ref{fig:qual_res_nuscenes}, and \emph{Virtual KITTI 2} dataset in figure \ref{fig:qual_res_vkitti}.

\section{Discussion and Future Work}
CubifAE-3D leverages the power of simulation to learn an embedding for RGB-to-depth, which is then used to train a network to predict object poses from RGB images on real datasets. We find that our method does surprisingly well on objects that are far from the ego-vehicle and is represented by only a few pixels, as is evident from our results on the KITTI \emph{moderate} and \emph{hard} difficulty levels. Additionally, with the \emph{cubification} technique, we are able to localize the position error to within each \emph{cuboid} so that it does not increase with increasing distance from the ego-camera. This encourages us to believe that with further quantification of the 3D space, we will be able to achieve higher accuracy and this will be explored in our future work. A non-uniform or a coarse-to-fine \emph{cubification} technique is also something of interest to fill the gap between short-range and long-range performance and reduce errors at cuboid boundaries. Furthermore, we would like to perform \emph{projective data augmentation} in order to virtually zoom and move the camera through space to increase the amount of labelled 3D pose training data so that the 3DOD network can generalize better on real datasets. 
% With the techniques presented in this paper, we already foresee new areas of research being explored. 

The techniques presented in this paper: pre-training an RGB-to-depth embedding from simulation and \emph{cubifying} 3D space are also likely to be useful for other AV perception tasks that are dependent on pixel-level depth embeddings, without the need for a ground-truth dense depth map, like free space/occupancy grid estimation or 6 DoF monocular localization of the ego-camera/AV.
% Using RGB-to-depth embeddings for performing a task that is highly dependent on understanding pixel-level depth without needing the ground-truth dense depth map is something that will open up the doors to many different ideas. 

We show that our method already outperforms some SoTA methods for monocular 3D object detection and we are hopeful that our continuing improvements to \emph{CubifAE-3D} will serve as a baseline for monocular 3D object detection.

\section{Conclusions}
3D object detection is a primary task in the perception pipeline for an AV and we present \emph{CubifAE-3D}, a method for doing this using monocular imagery. Our method utilizes dense depth information corresponding to RGB images from a Gaming Engine based simulator and this is used to pre-train a RGB-to-depth auto-encoder (AE). The encoder of this pre-trained AE is then detached and used as an embedder for RGB images. This embedding is subsequently used to train our 3D Object Detector (3DOD), which is trained in a fully supervised manner and requires RGB images and corresponding 3D object pose annotations. Our 3DOD \emph{cubifies} the 3D space around the camera into a grid of cuboids, where each cuboid is trained to output $N$ 3D object poses and their confidence scores. Once trained, our aggregate detector \emph{CubifAE-3D} (comprising of the AE-encoder and the 3DOD) is able to generate accurate 3D object poses and classifications from RGB images. The particular scheme of \emph{cubification} of the 3D space gives results whose accuracy is maintained with distance from the camera, with an average localization error of $0.2m$ and a never-exceed localization error of $0.4m$ upto a detection distance of $100m$ from the ego-vehicle for vehicle, pedestrian and cyclist classes in the KITTI dataset. We believe that \emph{CubifAE-3D} represents an important advancement in the field for fast, accurate 3D object detection from monocular images. The ability to do this robustly and accurately from cheap cameras (instead of expensive LIDAR) that are already present in a surround-view configuration in present generation commercial vehicles will be crucial in the development and democratization of AV technology.

\clearpage
{\small
\bibliographystyle{ieee_fullname}
\bibliography{main}
}

\clearpage
\setcounter{section}{0}
\renewcommand{\thesection}{S\arabic{section}}  
\setcounter{figure}{0}
\renewcommand{\thefigure}{S\alph{figure}}
\onecolumn
% This version of CVPR template is provided by Ming-Ming Cheng.
% Please leave an issue if you found a bug:
% https://github.com/MCG-NKU/CVPR_Template.

%%%%%%%%% TITLE
\begin{center}
\Large

\vspace{1.5in}

\textbf{Supplementary Materials - CubifAE-3D: Monocular Camera Space Cubification for Auto-Encoder based 3D Object Detection}

\vspace{0.5in}
\end{center}
\section{Supplementary Materials}

\subsection{Qualitative Results}

\begin{figure*}[h]
\centering
    \begin{subfigure}
    \centering
    \includegraphics[width=0.42\linewidth]{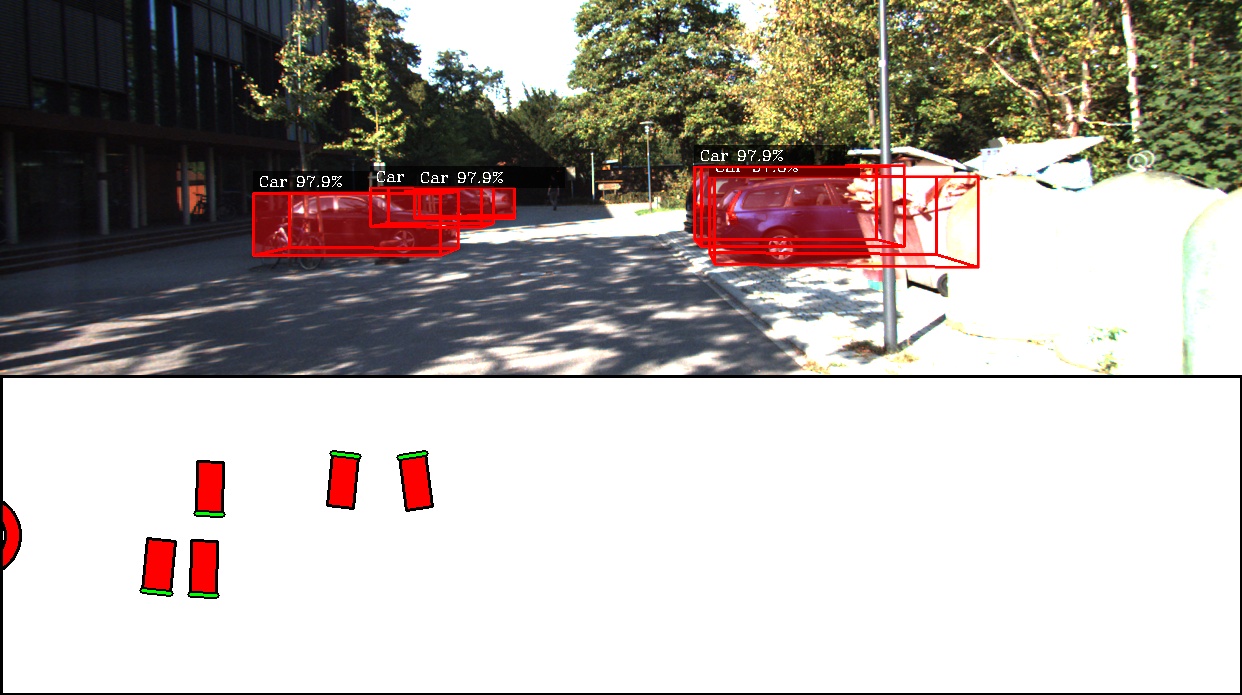}
    \end{subfigure}
    \hspace{0.05\textwidth}
    \begin{subfigure}
    \centering
    \includegraphics[width=0.42\linewidth]{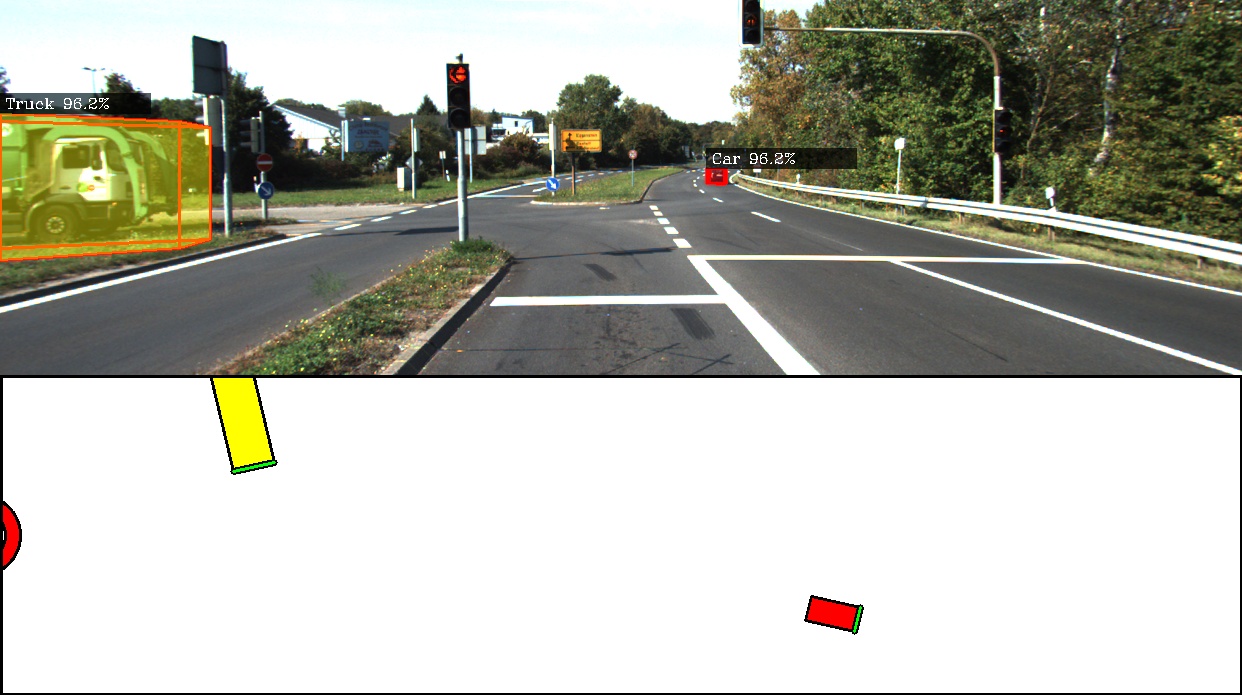}
    \end{subfigure}
    \hspace{0.05\textwidth}
    \begin{subfigure}
    \centering
    \includegraphics[width=0.42\linewidth]{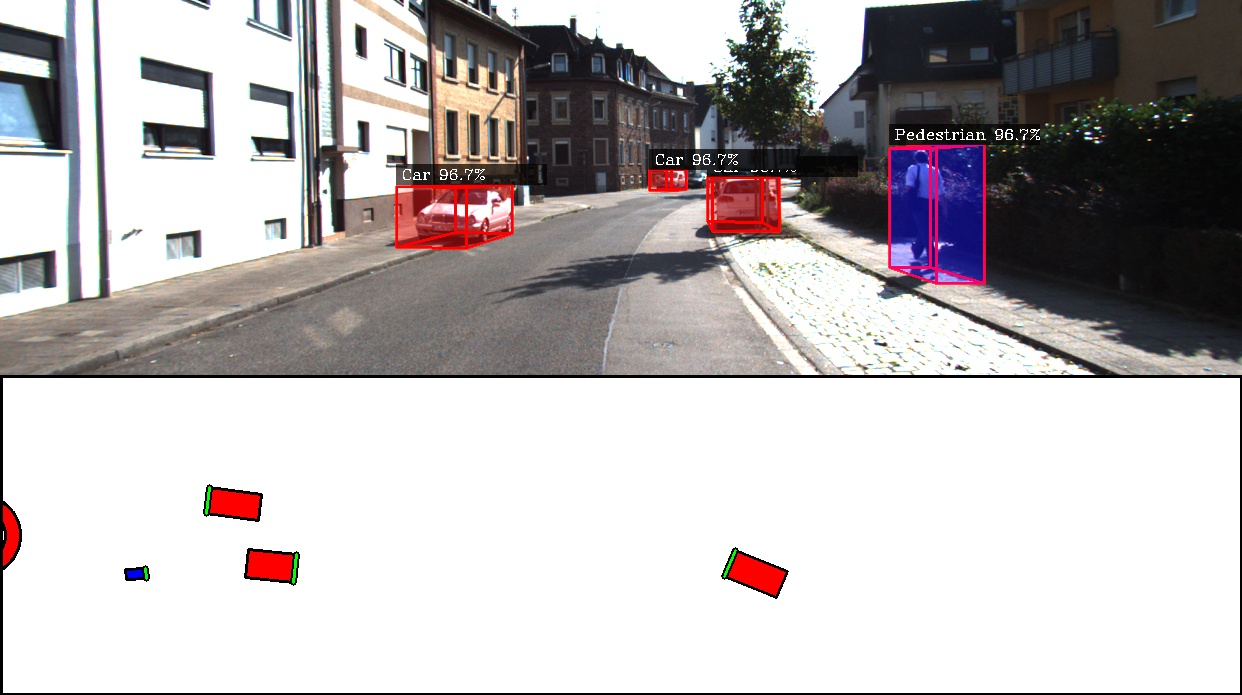}
    \end{subfigure}
    \hspace{0.05\textwidth}
    \begin{subfigure}
    \centering
    \includegraphics[width=0.42\linewidth]{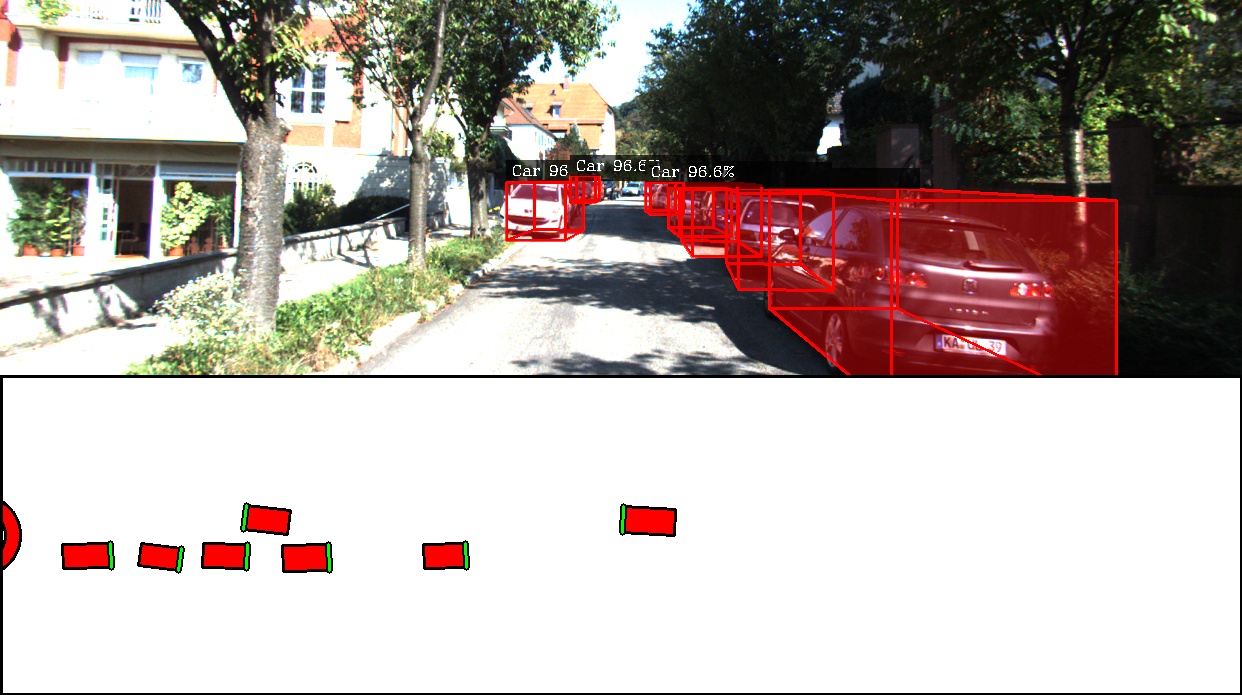}
    \end{subfigure}
    \hspace{0.05\textwidth}
    \begin{subfigure}
    \centering
    \includegraphics[width=0.42\linewidth]{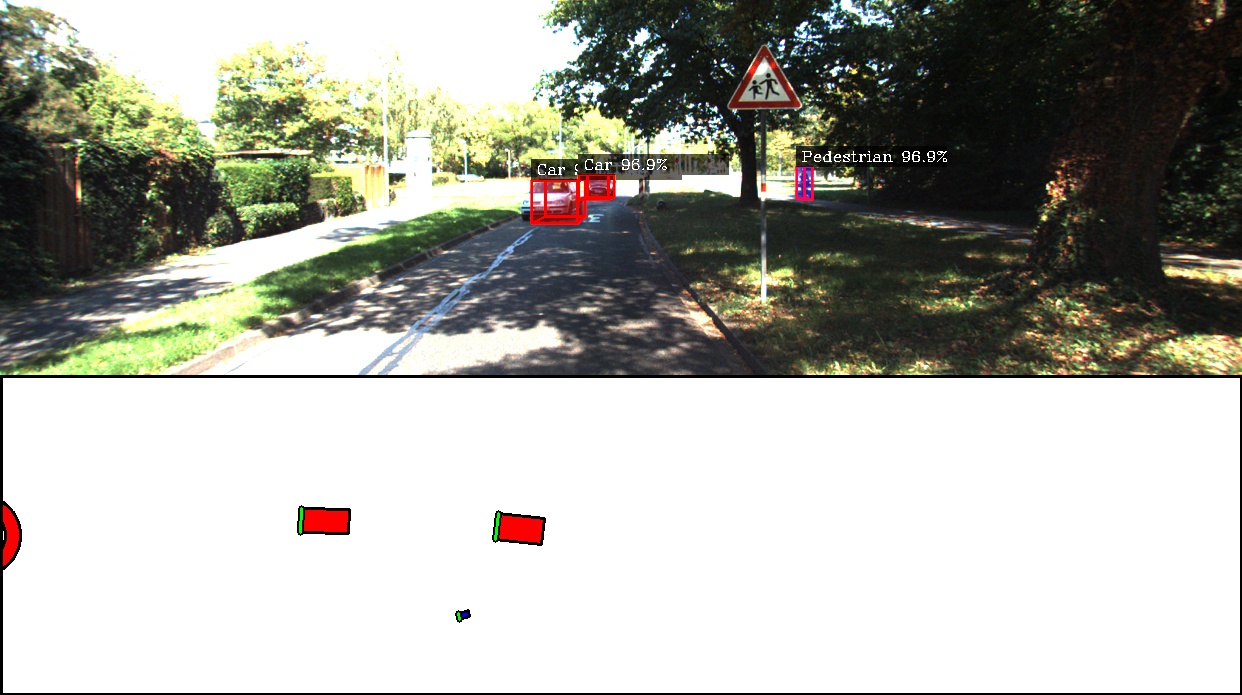}
    \end{subfigure}
    \hspace{0.05\textwidth}
    \begin{subfigure}
    \centering
    \includegraphics[width=0.42\linewidth]{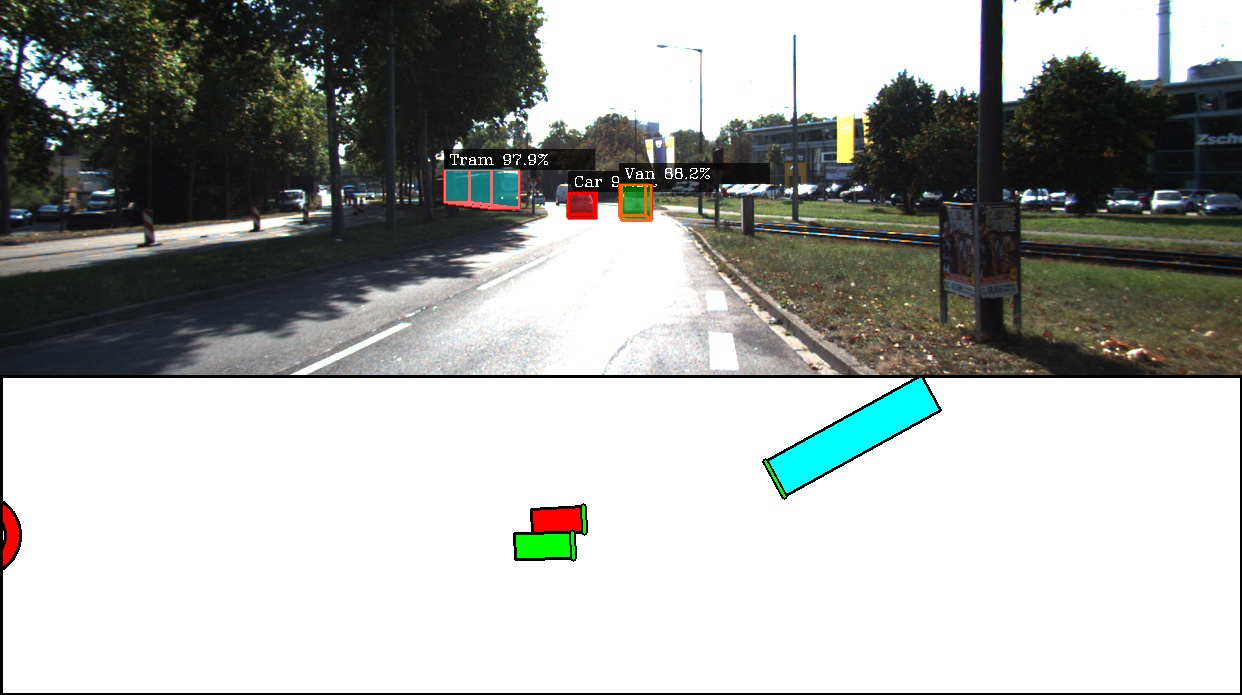}
    \end{subfigure}
\caption{Qualitative results on the \emph{KITTI} dataset. The top part of each image shows a bounding box obtained as a 2D projection of their 3D poses (\emph{red: car, yellow: truck, green: van, blue: pedestrian, cyan: tram, cyclist, and others}). The bottom part shows a birds-eye view of the object poses with the ego-vehicle positioned at the center of red circle drawn on the left; pointing towards the right of the image. 
\label{fig:qual_res_kitti}}
\end{figure*}

\begin{figure*}[h]
\centering
    \begin{subfigure}
    \centering
    \includegraphics[width=0.40\linewidth]{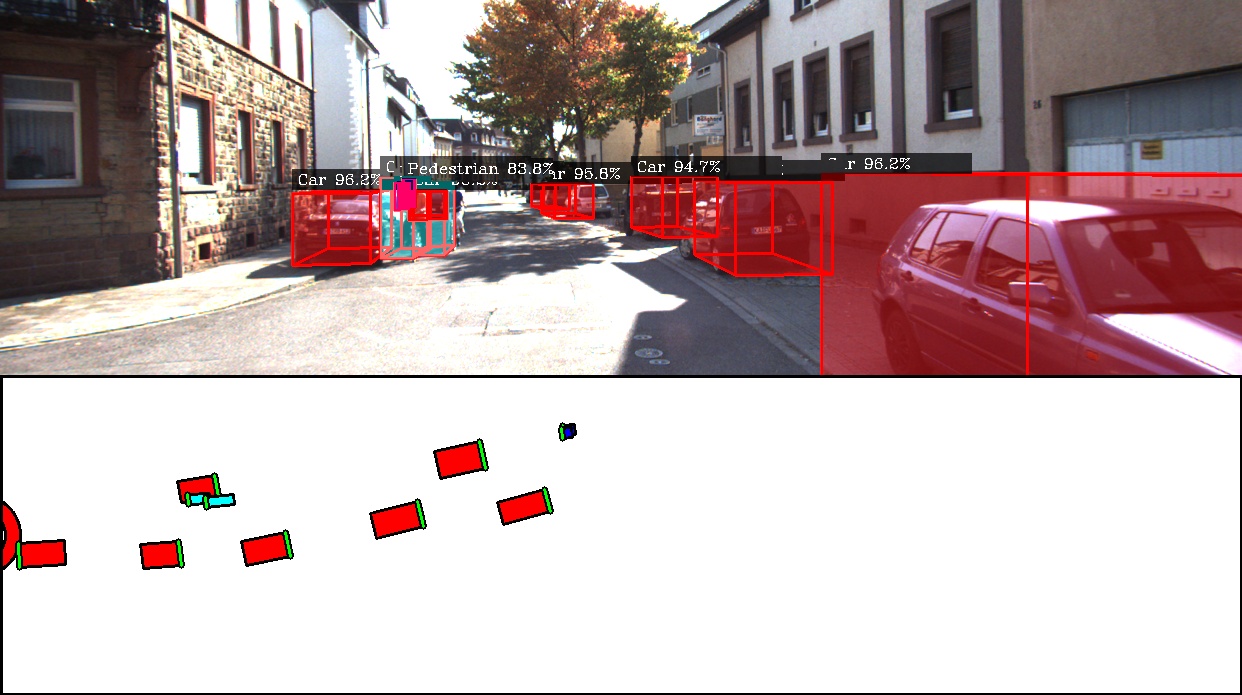}
    \end{subfigure}
    \hspace{0.05\textwidth}
    \begin{subfigure}
    \centering
    \includegraphics[width=0.40\linewidth]{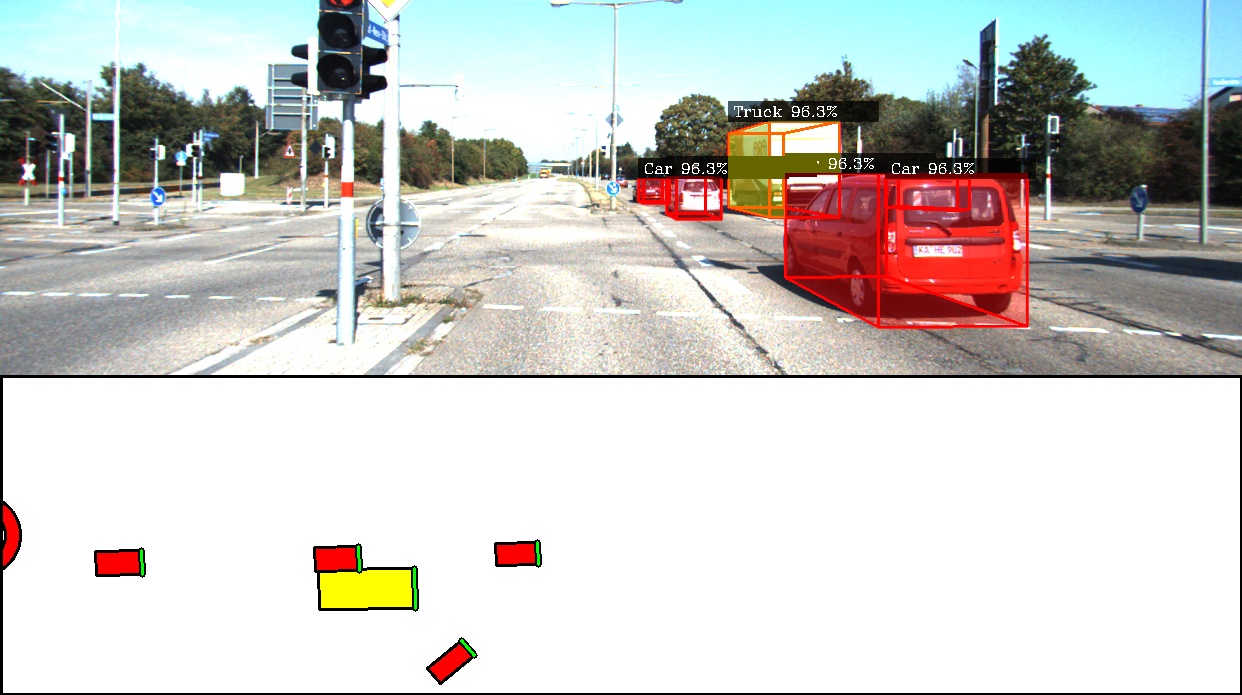}
    \end{subfigure}
    \hspace{0.05\textwidth}
    \begin{subfigure}
    \centering
    \includegraphics[width=0.40\linewidth]{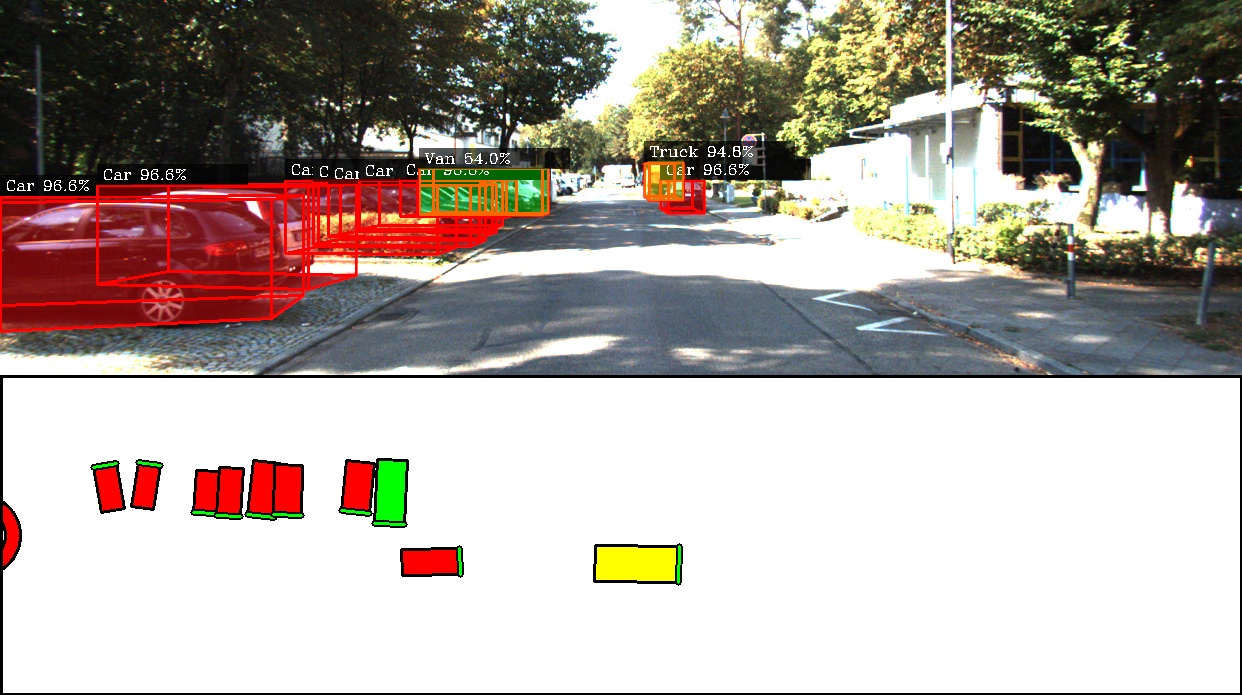}
    \end{subfigure}
    \hspace{0.05\textwidth}
    \begin{subfigure}
    \centering
    \includegraphics[width=0.40\linewidth]{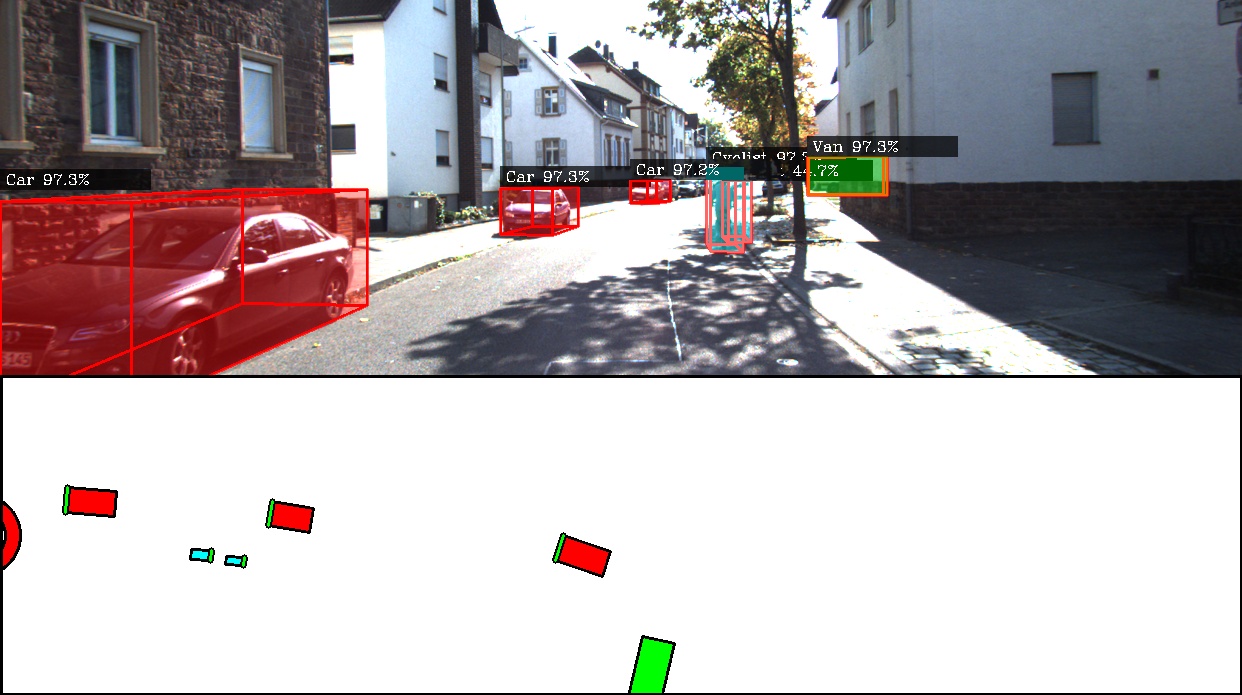}
    \end{subfigure}
    \hspace{0.05\textwidth}
    \begin{subfigure}
    \centering
    \includegraphics[width=0.40\linewidth]{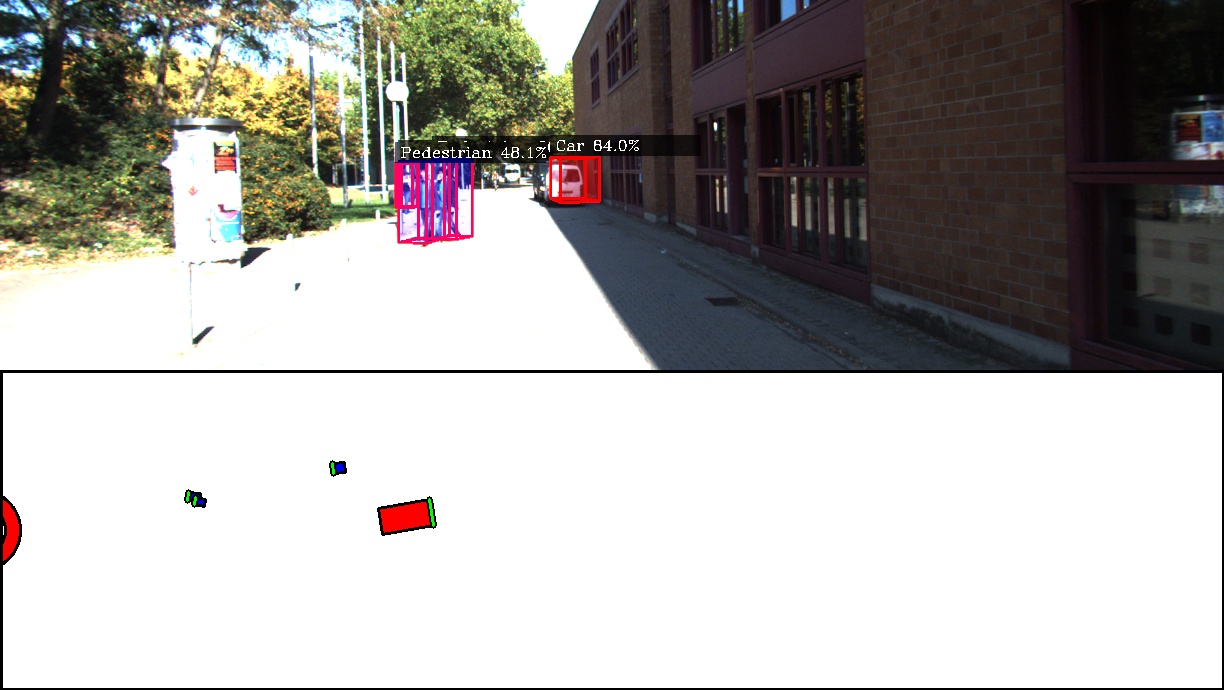}
    \end{subfigure}
    \hspace{0.05\textwidth}
    \begin{subfigure}
    \centering
    \includegraphics[width=0.40\linewidth]{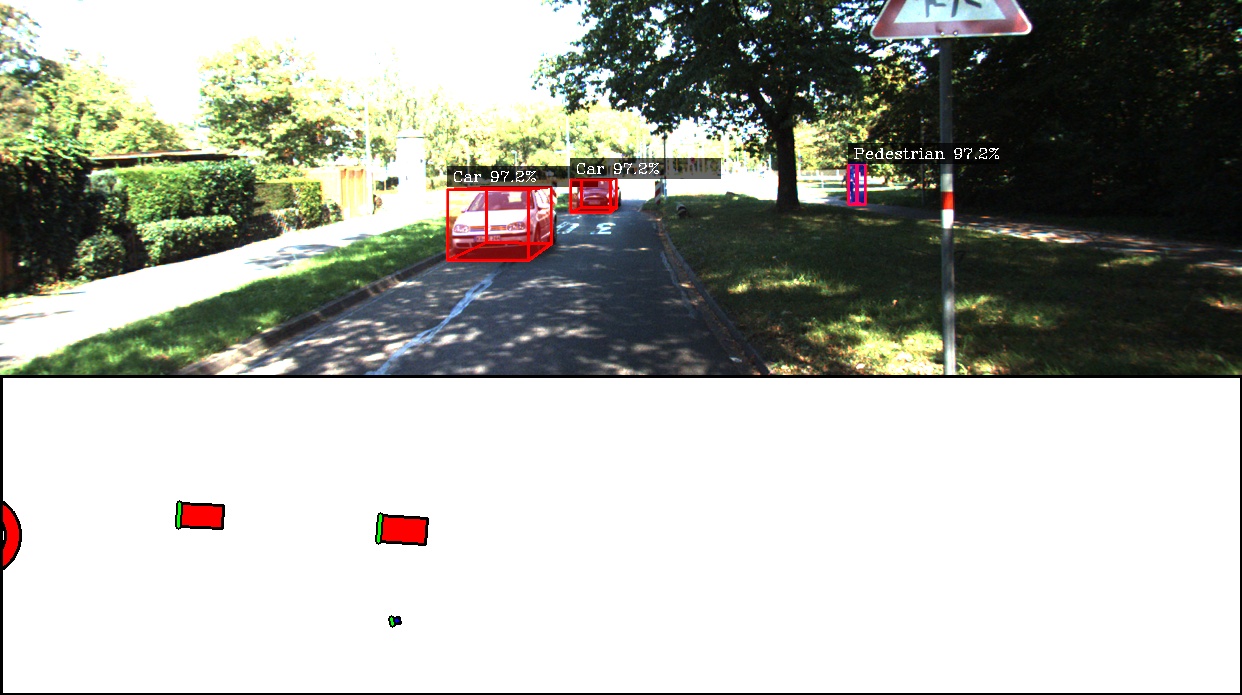}
    \end{subfigure}
    \hspace{0.05\textwidth}
    \begin{subfigure}
    \centering
    \includegraphics[width=0.40\linewidth]{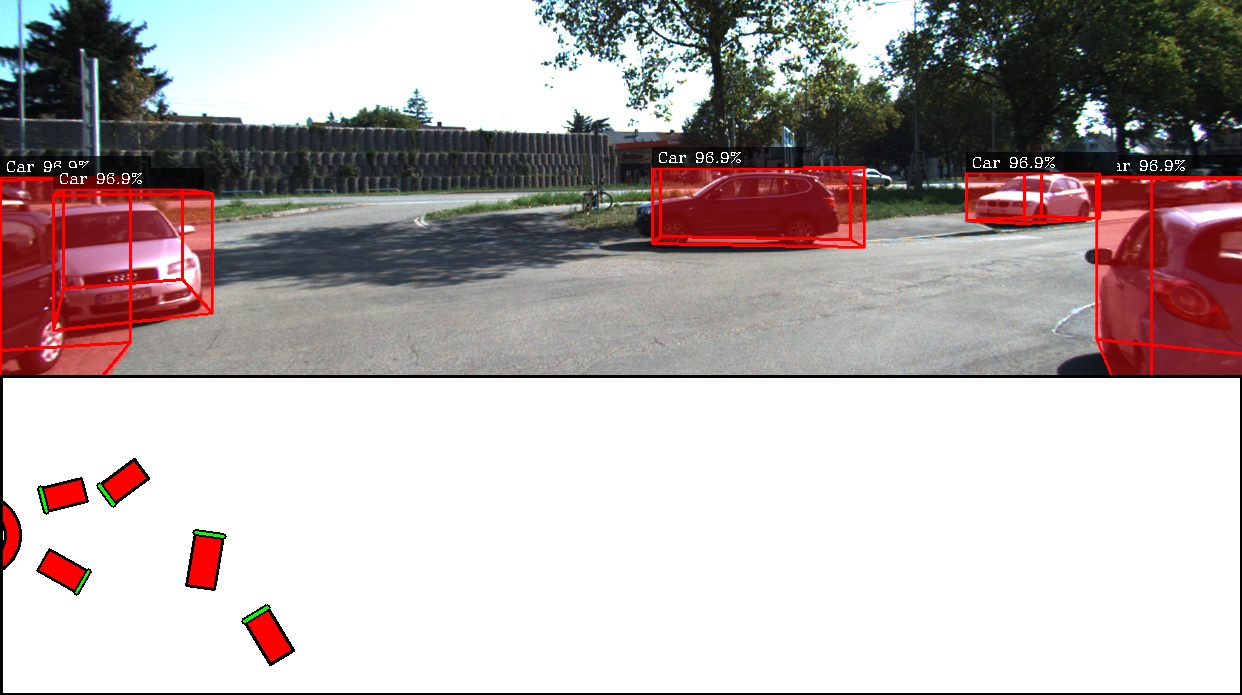}
    \end{subfigure}
    \hspace{0.05\textwidth}
    \begin{subfigure}
    \centering
    \includegraphics[width=0.40\linewidth]{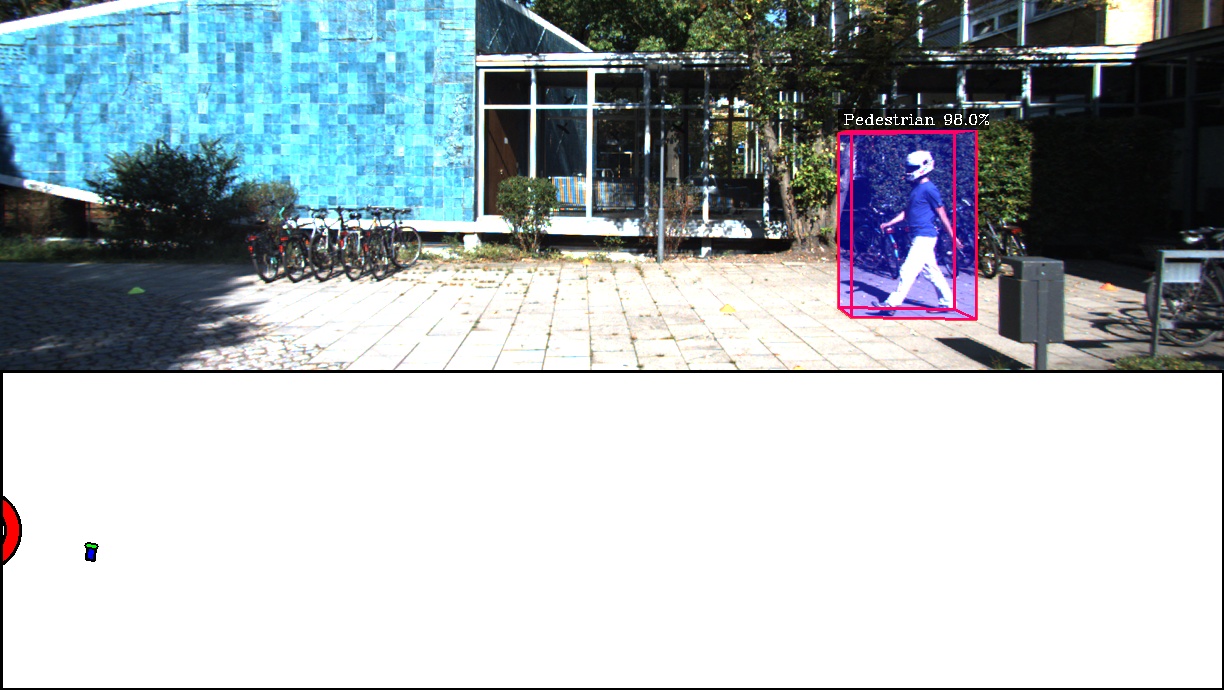}
    \end{subfigure}
    \hspace{0.05\textwidth}
    \begin{subfigure}
    \centering
    \includegraphics[width=0.40\linewidth]{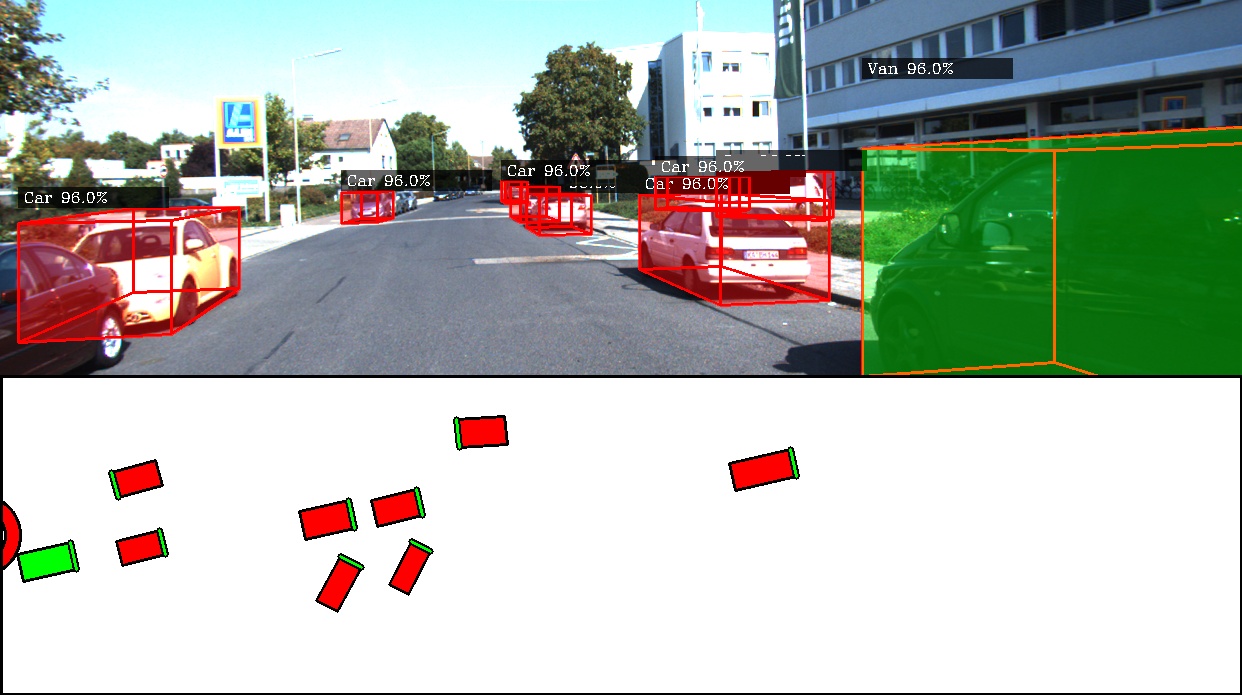}
    \end{subfigure}
    \hspace{0.05\textwidth}
    \begin{subfigure}
    \centering
    \includegraphics[width=0.40\linewidth]{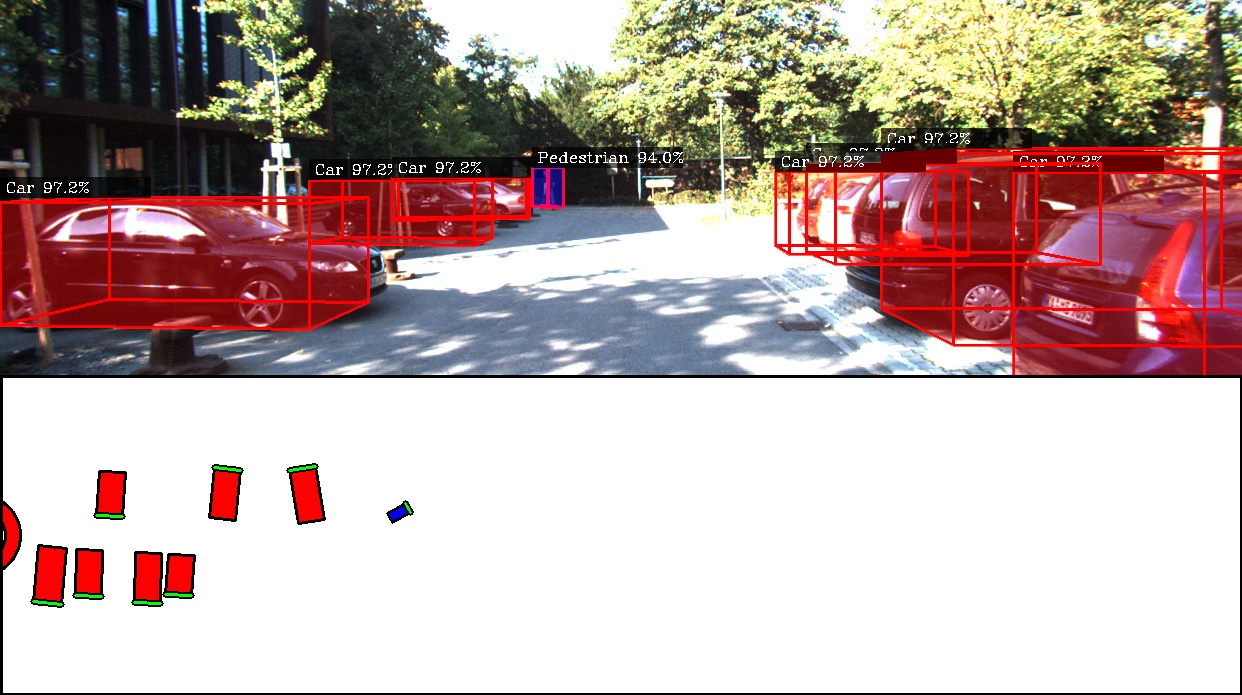}
    \end{subfigure}
\caption{Qualitative results on the \emph{KITTI} dataset. The top part of each image shows a bounding box obtained as a 2D projection of their 3D poses (\emph{red: car, yellow: truck, green: van, blue: pedestrian, cyan: tram, cyclist, and others}). The bottom part shows a birds-eye view of the object poses with the ego-vehicle positioned at the center of red circle drawn on the left; pointing towards the right of the image. 
\label{fig:qual_res_kitti_2}}
\end{figure*}

\begin{figure*}[h]
\centering
    \begin{subfigure}
    \centering
    \includegraphics[width=0.40\linewidth]{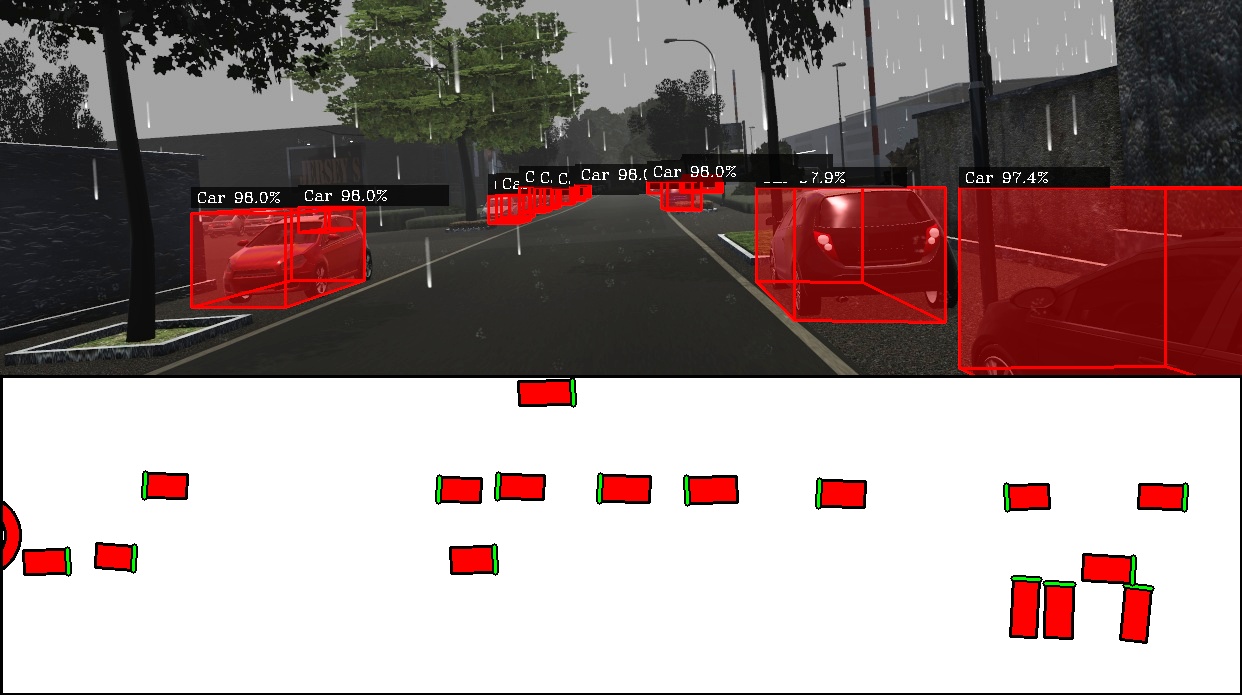}
    \end{subfigure}
    \hspace{0.05\textwidth}
    \begin{subfigure}
    \centering
    \includegraphics[width=0.40\linewidth]{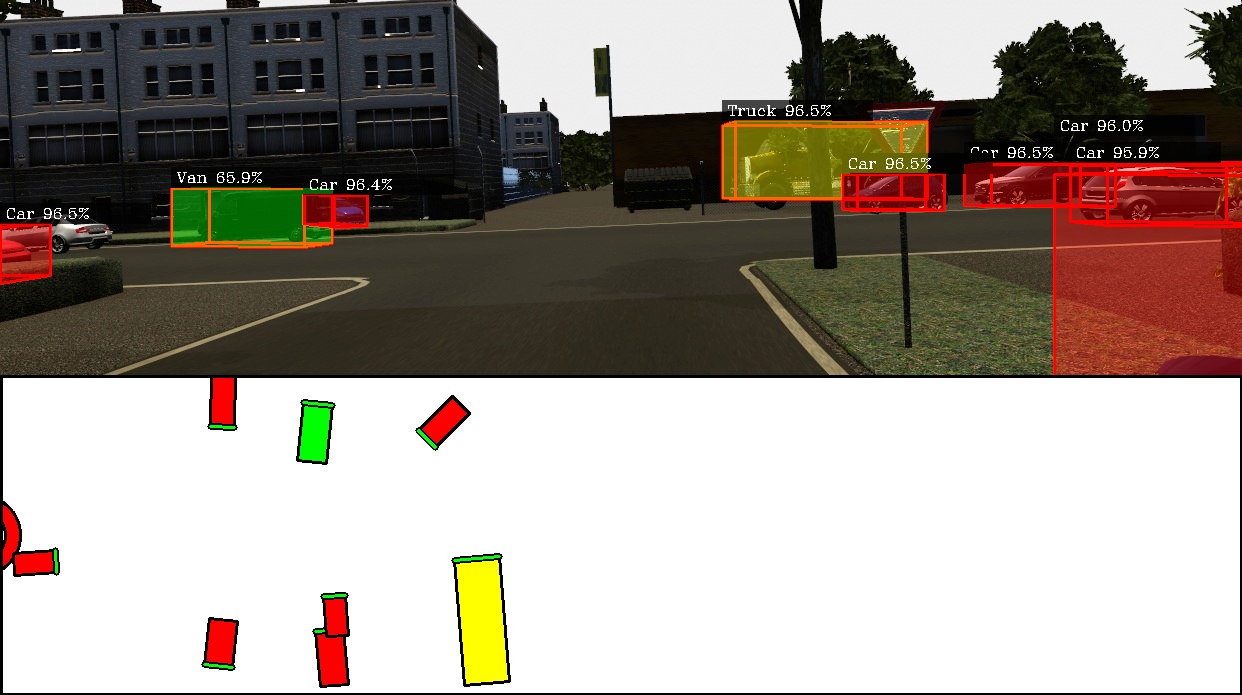}
    \end{subfigure}
    \hspace{0.05\textwidth}
    \begin{subfigure}
    \centering
    \includegraphics[width=0.40\linewidth]{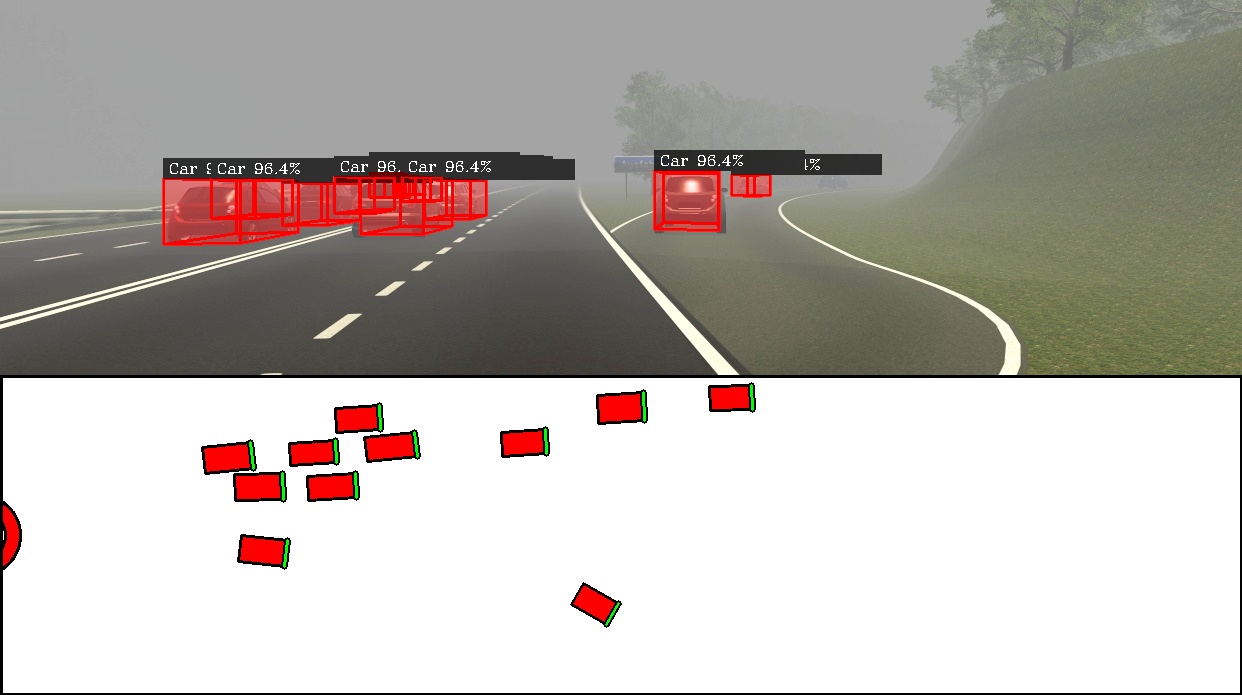}
    \end{subfigure}
    \hspace{0.05\textwidth}
    \begin{subfigure}
    \centering
    \includegraphics[width=0.40\linewidth]{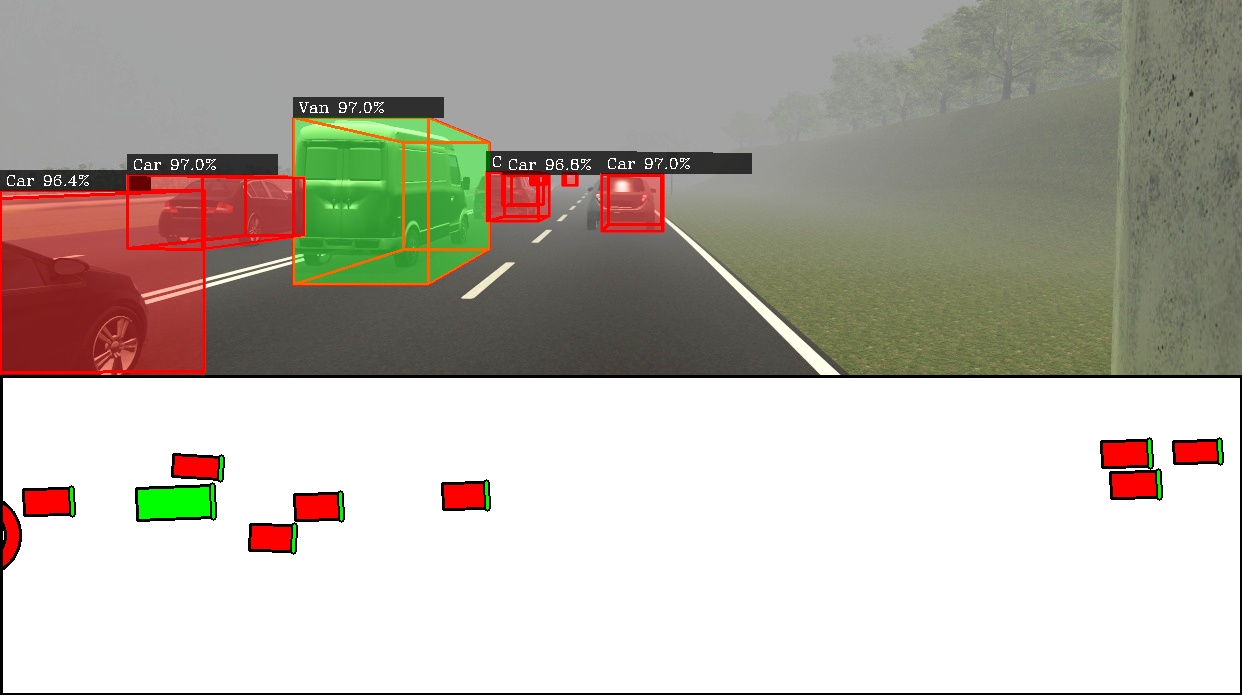}
    \end{subfigure}
    \hspace{0.05\textwidth}
    \begin{subfigure}
    \centering
    \includegraphics[width=0.40\linewidth]{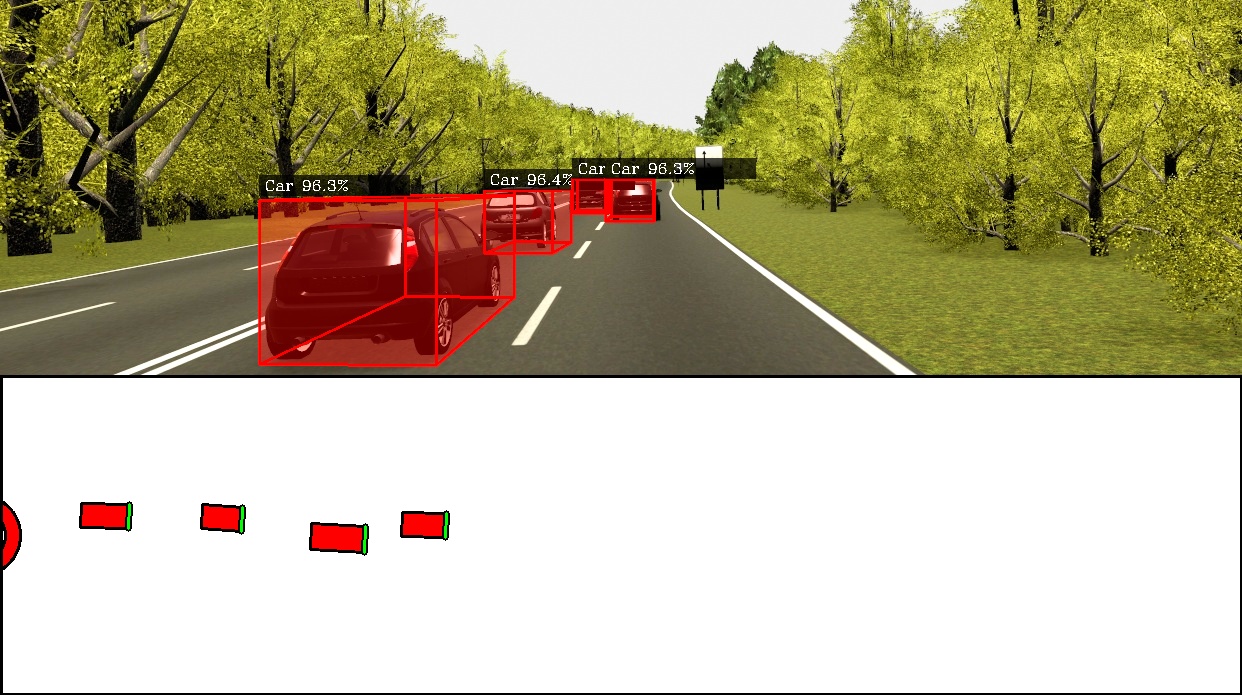}
    \end{subfigure}
    \hspace{0.05\textwidth}
    \begin{subfigure}
    \centering
    \includegraphics[width=0.40\linewidth]{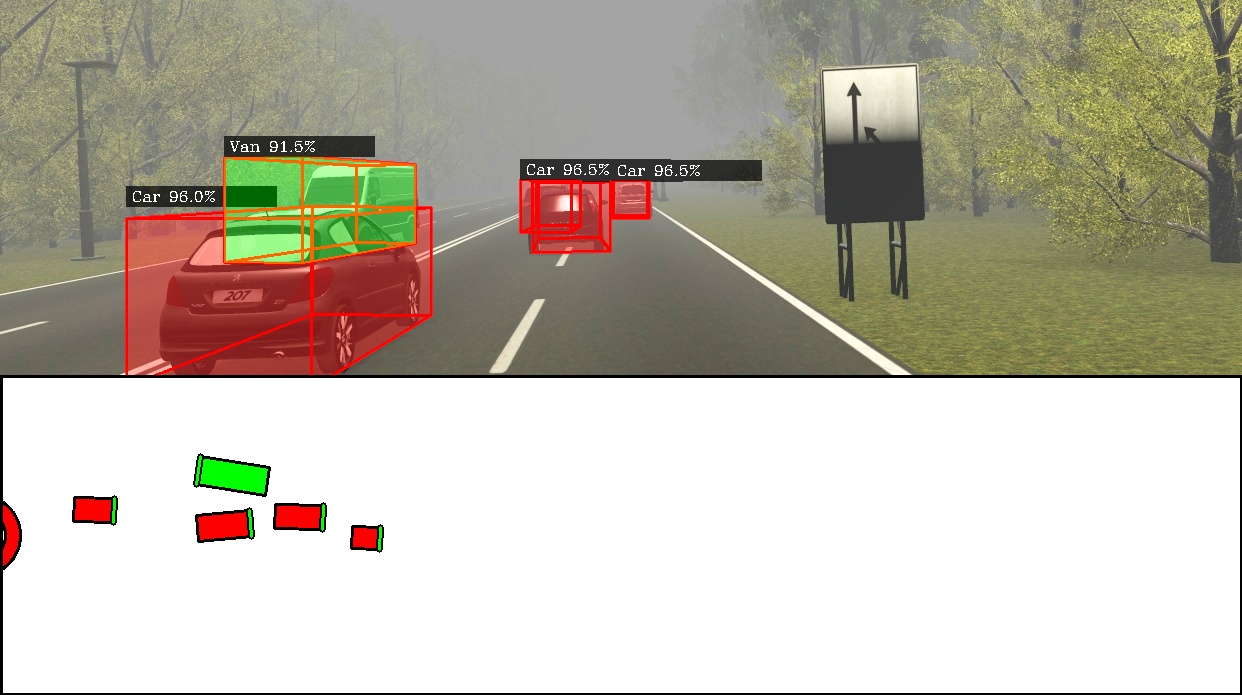}
    \end{subfigure}
    \hspace{0.05\textwidth}
    \begin{subfigure}
    \centering
    \includegraphics[width=0.40\linewidth]{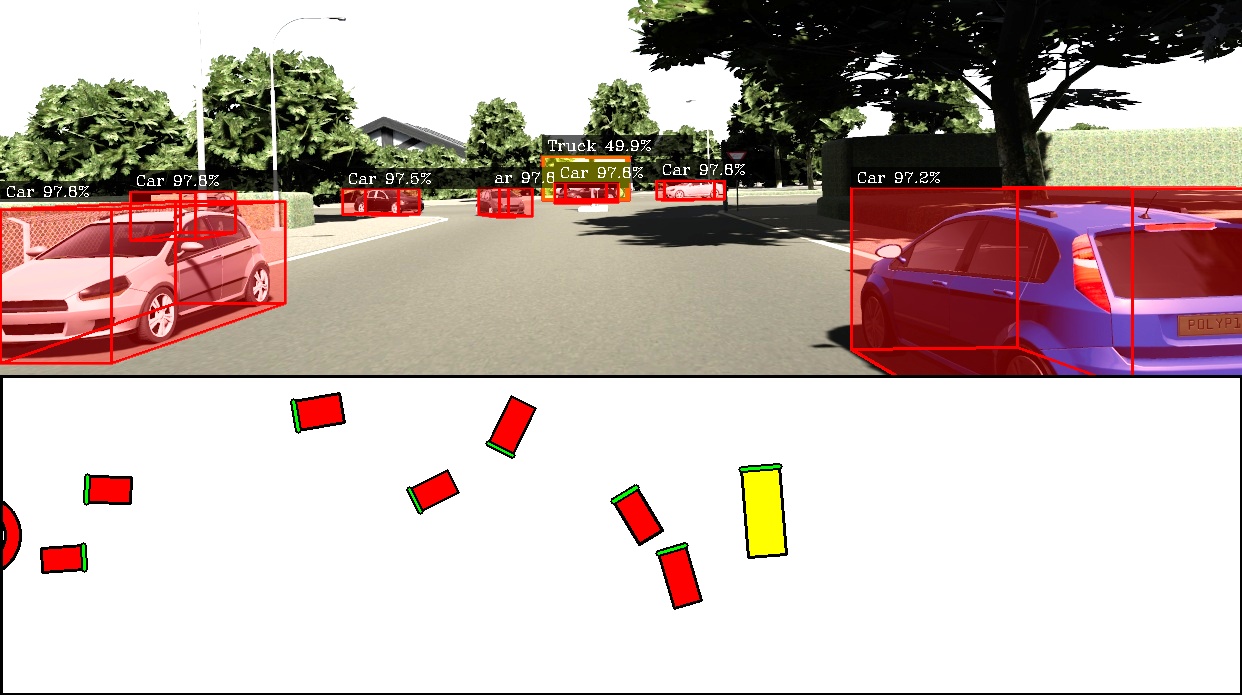}
    \end{subfigure}
    \hspace{0.05\textwidth}
    \begin{subfigure}
    \centering
    \includegraphics[width=0.40\linewidth]{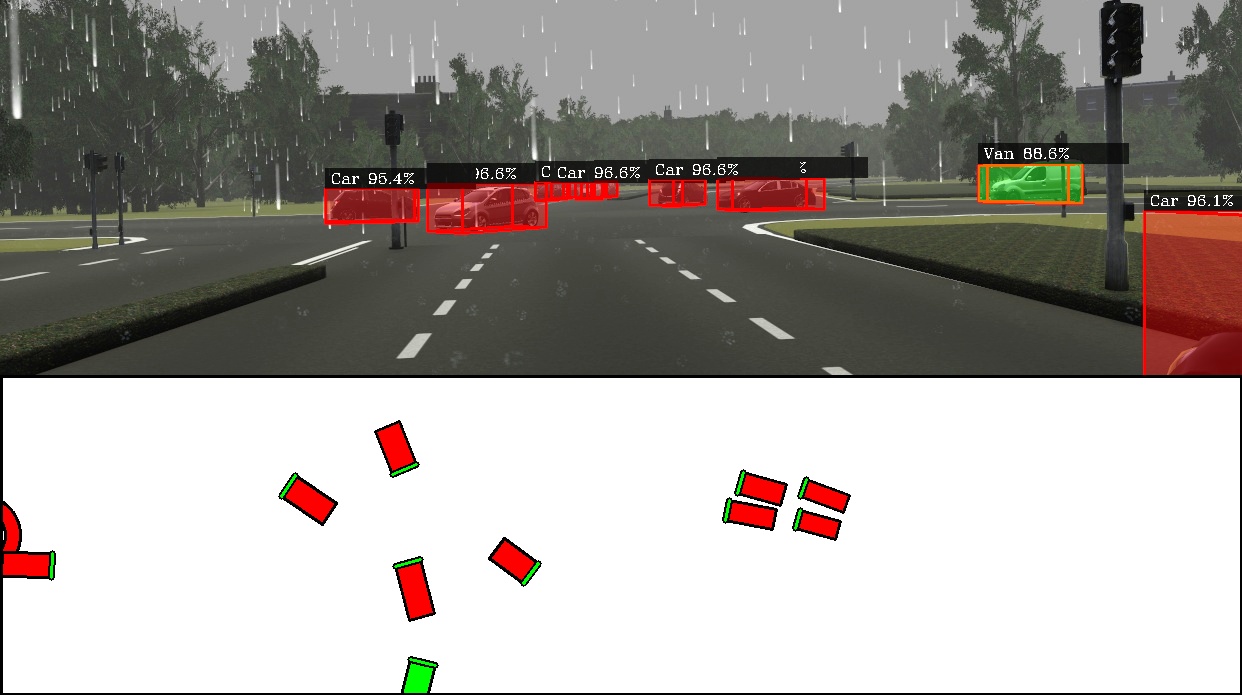}
    \end{subfigure}
    \hspace{0.05\textwidth}
    \begin{subfigure}
    \centering
    \includegraphics[width=0.40\linewidth]{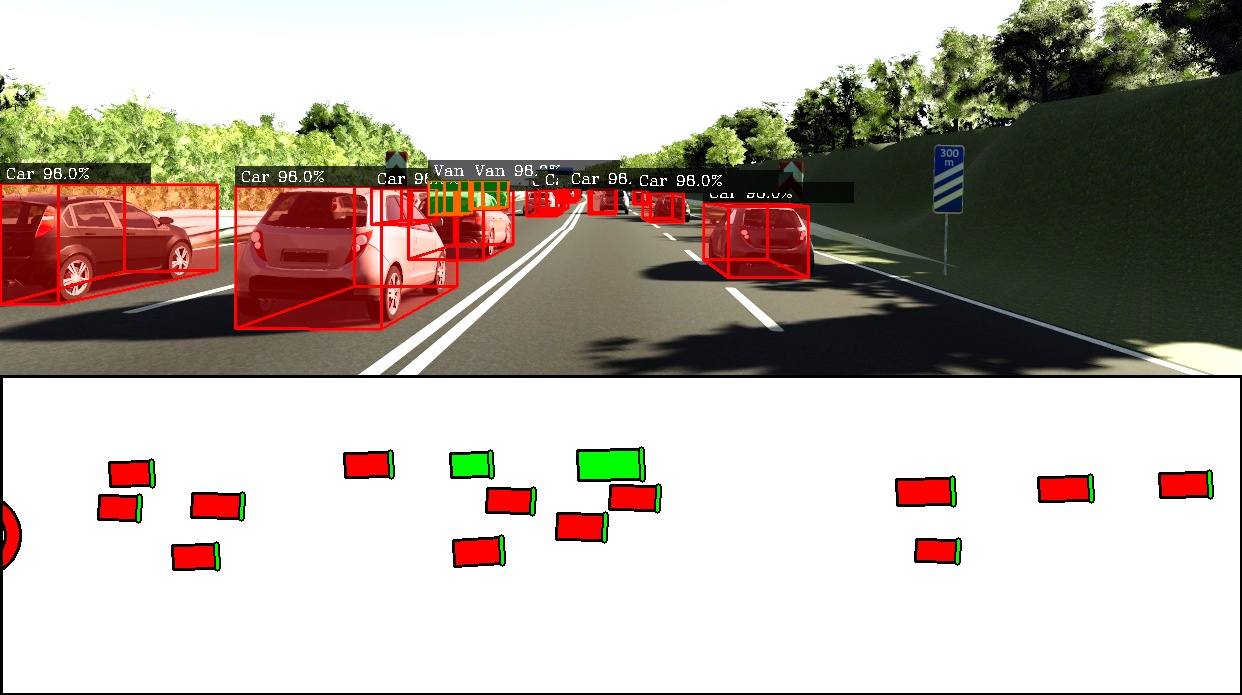}
    \end{subfigure}
    \hspace{0.05\textwidth}
    \begin{subfigure}
    \centering
    \includegraphics[width=0.40\linewidth]{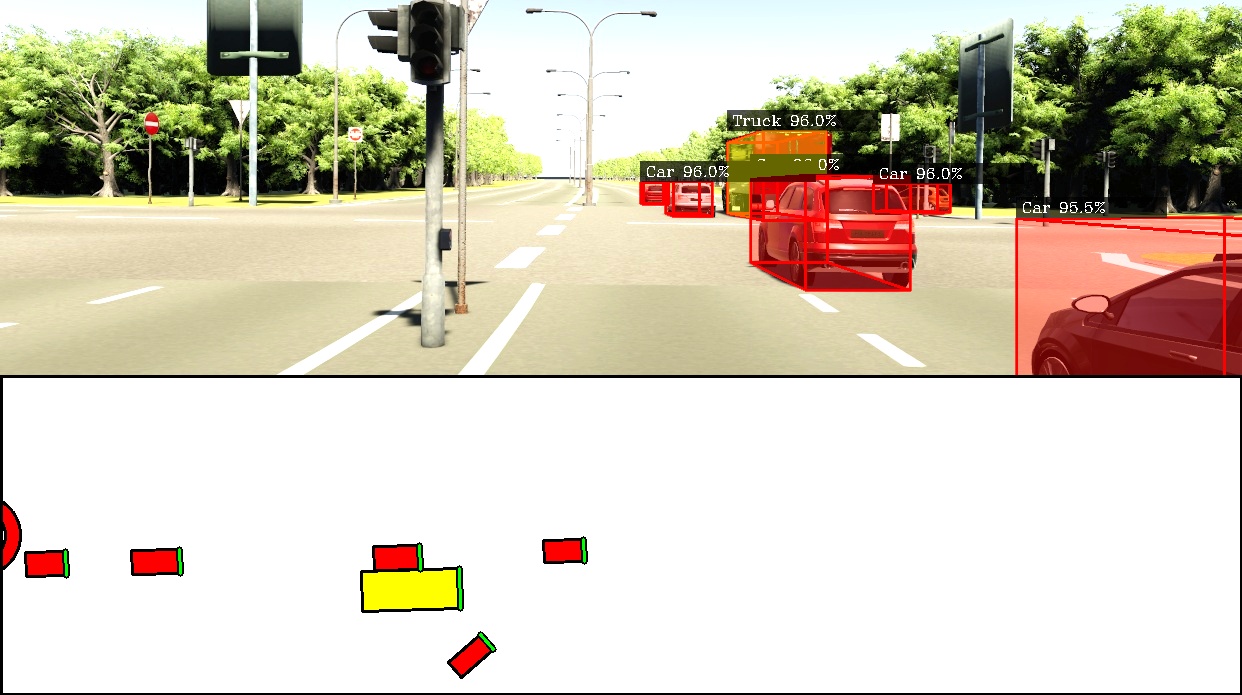}
    \end{subfigure}
\caption{Qualitative results on the \emph{Virtual KITTI 2} dataset across various scenes, weather, and lighting conditions. The top part of each image shows a bounding box obtained as a 2D projection of their 3D poses (\emph{red: car, yellow: truck, cyan: van}). The bottom part shows a birds-eye view of the object poses with the ego-vehicle positioned at the center of red circle drawn on the left; pointing towards the right of the image. 
\label{fig:qual_res_vkitti}}
\end{figure*}

This section demonstrates qualitative results of our method on KITTI object detection dataset (\textit{val} split) [figure \ref{fig:qual_res_kitti}, figure \ref{fig:qual_res_kitti_2}] and Virtual KITTI 2 dataset [figure \ref{fig:qual_res_vkitti}]. Number of classes considered for the former is $8$ and for the latter is $3$. Different classes are colored differently, more details in each figure caption. 
\vspace{0.5in}

\end{document}